\documentclass{article}

\usepackage[preprint, final, nonatbib]{neurips_2026}

\usepackage[utf8]{inputenc} % allow utf-8 input
\usepackage[T1]{fontenc}    % use 8-bit T1 fonts
\usepackage{hyperref}       % hyperlinks
\usepackage{url}            % simple URL typesetting
\usepackage{booktabs}       % professional-quality tables
\usepackage{amsfonts}       % blackboard math symbols
\usepackage{nicefrac}       % compact symbols for 1/2, etc.
\usepackage{microtype}      % microtypography
\usepackage{xcolor}         % colors

\usepackage{graphicx}
\usepackage{amsmath}
\usepackage{amssymb}
\usepackage[capitalize, nameinlink]{cleveref}
\usepackage{braket}
\usepackage{algorithm}
\usepackage{algpseudocode}
\newtheorem{definition}{Definition}
\newtheorem{theorem}{Theorem}

\usepackage[giveninits=true,sorting=none,citestyle=numeric-comp]{biblatex}
\newcommand{\deriv}[1]{\frac{\mathrm{d}}{\mathrm{d}#1}}
\newcommand{\integral}[4]{\int_{#1}^{#2} #3\ \mathrm{d}#4}

\newcommand{\pderivs}[2]{\frac{\partial #1}{\partial #2}}

\newcommand{\heaviside}[1]{\Theta\left( #1 \right)}
\newcommand{\taus}{\tau_\mathrm{s}}
\newcommand{\taum}{\tau_\mathrm{m}}
\newcommand{\thresh}{\vartheta}
\newcommand{\outspikei}[1]{t_{#1}}
\newcommand{\outspike}[1]{\pmb t}
\newcommand{\causal}{\mathcal{C}}
\newcommand{\presyn}{\mathcal{K}}
\newcommand{\paths}{\mathcal{P}}
\newcommand{\weightsumdraft}[3]{\sum_{#2 \in #3_{#1}} W_{#1#2}}
\newcommand{\weightsum}[2]{\weightsumdraft{#1}{#2}{\causal}}

\newcommand{\weightsumth}[2]{\weightsum{#1}{#2} - \thresh}

\newcommand{\expsumT}[2]{\sum_{#2 \in \causal_{#1}} W_{#1#2} \subs{#2}}

\newcommand{\subs}[1]{T_{#1}}

\usepackage{tcolorbox} 
\definecolor{challengegrey}{RGB}{240, 240, 240} % light neutral grey
\tcbset{
  challengebox/.style={
  colback=challengegrey,
  colframe=black!30,
  coltitle=black,
  boxrule=0pt,
  arc=4pt,
  left=0pt,
  right=0pt,
  top=0pt,
  bottom=0pt,
  fonttitle=\bfseries
}}

\newtcolorbox{challengebox}{
  colback=gray!10,
  colframe=gray!25,
  left=0pt,
  right=0pt,
  top=-1.5pt,
  bottom=-1.5pt,
  boxrule=0.pt,
  arc=4pt,
}

\title{Causal pieces: analysing and improving spiking neural networks piece by piece}

\author{%
  Dominik Dold,\ Philipp Christian Petersen \\
  Faculty of Mathematics and Research Network Data Science @ Uni Vienna\\
  University of Vienna\\
  Kolingasse 14-16, 1090 Vienna, Austria \\
}

\begin{document}

\maketitle

\begin{abstract}
We introduce \textit{causal pieces}, a novel concept for analysing spiking neural networks (SNNs), inspired by \textit{linear pieces} used to study expressivity and trainability in artificial neural networks (ANNs). 
Causal pieces partition the input and parameter space of a feedforward SNN with single-spike coding into distinct regions where the same subnetwork causes the output spikes.
For networks of current-based leaky integrate-and-fire (LIF) neurons with large membrane time constants, we show that within each causal piece, output spike times are locally Lipschitz continuous with respect to inputs and network parameters. We further prove a lower bound on the approximation error that depends on the number of causal pieces.
Thus, the number of causal pieces is a measure of the approximation capabilities of SNNs, which is valid despite spike-time discontinuities and applies to networks with both excitatory and inhibitory synapses. 
Empirically, we find that parameter initialisations yielding more causal pieces on the training set strongly correlate with SNN training success across multiple benchmarks, including Yin-Yang, Fashion-MNIST, and EuroSAT. Moreover, simulations with standard single-spike LIF neurons indicate that our findings extend beyond the theoretically analysed regime. 
These results establish causal pieces as a powerful and principled tool for analysing and improving the computational capabilities of SNNs.

\end{abstract}

\section{Introduction}
Spiking neural networks (SNNs) have recently received increased attention due to their ability to facilitate low-power hardware solutions for deep learning methods, particularly for edge applications, e.g., in outer space onboard spacecraft \cite{izzo2022neuromorphic,schumann_radiation_2022,lunghi2024investigation}.
In large parts, this is caused by the development of methods and software tools that allow the use of error backpropagation to train SNNs \cite{neftci2019surrogate,mostafa2017supervised,goltz2021fast,comsa2020temporal,klos2025smooth}, as well as emerging spike-based hardware systems \cite{frenkel2023bottom} such as Intel's digital Loihi \cite{davies_loihi_2018,orchard2021efficient} and the analog BrainScaleS-2 \cite{cramer2022surrogate,spilger2023hxtorch} chip.
Despite these developments, the computational capabilities of SNNs remain poorly understood \cite{singh2023expressivity,neuman2024stable}, and it is still an open question whether spike-based models provide advantages compared to non-spiking artificial neural networks (ANNs) \cite{davidson2021comparison,yin2021accurate,kucik2021investigating,lunghi2024investigation,dampfhoffer2022snns}.

Existing theoretical analyses of SNNs are often limited in scope.
Early theoretical results based on VC dimension are limited to binary classification tasks and abstract spiking neuron models \cite{maass1997networks,maass1999complexity,zador1996vc}.
More recent approaches introduce additional assumptions to increase analytical tractability and avoid spike-time discontinuities, such as single-spike coding, positive weights, feedforward architectures, and simplified membrane and synapse dynamics (e.g. linearly rising postsynaptic potentials) \cite{stanojevic2023exact,neuman2024stable,singh2023expressivity}. 
Generally, the fact that output spike times can change discontinuously or vanish due to continuous changes in inputs or parameters has so far prevented a rigorous analysis of SNN expressivity.
% Thus, a general framework for analysing SNNs without relying on such restrictive assumptions remains missing.

To approach this challenge, we introduce a framework for characterising SNNs inspired by \textit{linear pieces}, a concept used to analyse the expressivity of feedforward ReLU-based ANNs \cite{frenzen2010number,montufar2014number,telgarsky2016benefits,hanin2019complexity}.
Our main contribution is the idea of \textit{causal pieces}: a structured way to partition the input and parameter space of a feedforward SNN with single-spike coding into regions where output spikes are caused by the same subnetwork.
Causal pieces naturally account for discontinuities in spike timing, can be evaluated for commonly used spiking neuron models, and do not require restrictions on synapse types.
While piecewise decompositions in discrete-time spiking models \cite{nguyen2025time} have recently been introduced,  these differ fundamentally from our approach: their regions correspond to constant function outputs, and their number does not increase with neural network depth. 

\begin{figure*}[t!]
    \centering
\includegraphics[width=.805\textwidth]{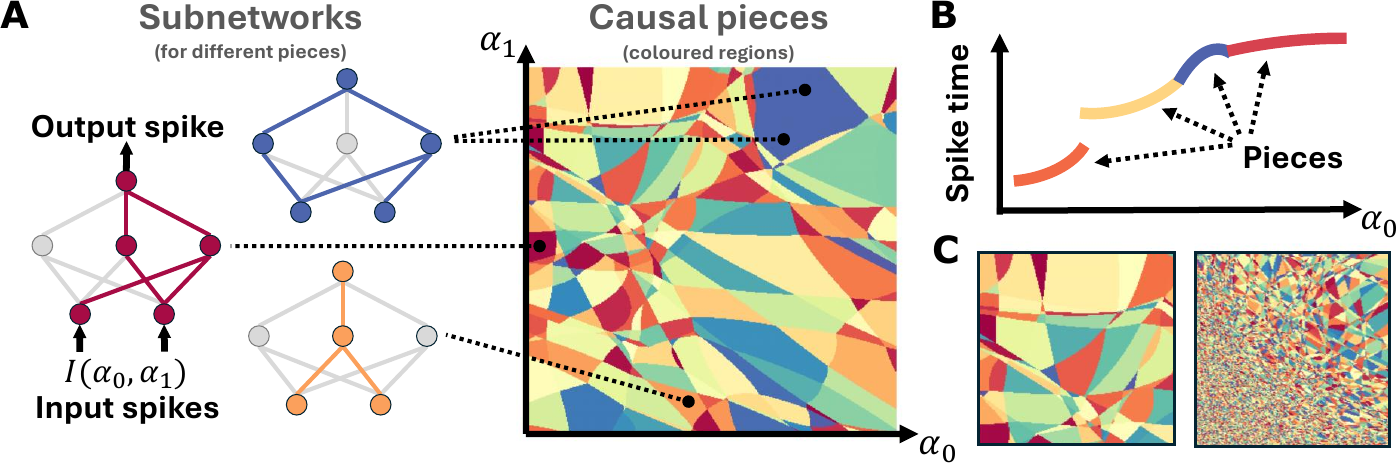}
    % \vspace{-5mm}
	\caption{Causal pieces.
    \textbf{(A)} Every coloured region is a causal piece, fully partitioning the input space (here: 2D inputs). Within each piece, the same subnetwork causes the SNN output (left, coloured like the corresponding piece). Here we scan a 2D hyperplane $I$ in the $40$-dimensional input space, parametrised by $(\alpha_0, \alpha_1) \in [0,1]^2$.
    \textbf{(B)} On each piece (coloured), the output spike time of the SNN is a continuous function.
    \textbf{(C)} Causal pieces for an SNN with no (left) and one (right) hidden layer.}
	\label{fig:Intro}
\end{figure*}

\cref{fig:Intro} illustrates our concept.
Each causal piece corresponds to a region of the inputs and network parameters in which the same subset of neurons and synapses caused the output spikes (\cref{fig:Intro}A). 
Within a piece, the output spike times vary smoothly, while transitions between pieces can induce discontinuities (\cref{fig:Intro}B). 
Although the number of pieces is determined by the network parameters, it can be vastly increased by adding more layers (\cref{fig:Intro}C, shown architectures: $[40,1]$, $[40,40,1]$).

This work establishes causal pieces as a tool for analysing feedforward SNNs both theoretically and empirically.
We show that the number of causal pieces provides a measure of expressivity by deriving a lower bound on the approximation error for single-spike, current-based leaky integrate-and-fire (LIF) neurons with large membrane time constant (\cref{theorem:nLifbound}).
Furthermore, we demonstrate that the number of causal pieces the training data fall into at network initialisation correlates with training success (\cref{fig:PiecesVsAcc,fig:fashion}), providing a grounded metric to guide SNN initialisation currently missing in the literature \cite{rossbroich2022fluctuation}. 
Notably, even SNNs with only positive weights can realise a large number of causal pieces and achieve competitive performance compared to fully connected ANNs.
While our theoretical results are derived for a specific class of LIF neurons, simulations with standard single-spike LIF neurons indicate that key results may extend to more complex neuron models.
Moreover, our derived bounds generalise to multi-spike neurons via a recently introduced transference principle \cite{dold2026equivalence}, which also provides a way of extending the definition of causal pieces to the multi-spike setting.

In the following, we briefly introduce the preliminaries required to follow this study before providing theoretical and experimental results. Mathematical proofs, simulation details, and algorithms can be found in the Appendix. 
Code is publicly available on GitHub \cite{dold2026snnpiece}.

\section{Methods}\label{sec:method}
In this work, we focus on the widely used current-based leaky integrate-and-fire neuron model in the limit of very large membrane time constants $\taum$ (see \cref{def:nLIF}), which is also known as the nLIF model \cite{mostafa2017supervised,goltz2021fast}.
While this type of neuron model features no leakage, input spikes are still integrated with a finite synaptic time constant $\taus$.
In addition, we assume a widely used purely time-dependent encoding scheme in which each neuron spikes at most once \cite{comsa2020temporal,goltz2021fast,stanojevic2023exact,goltz2024delgrad,che2024ettfs,klos2025smooth}, which is also known as phase, latency, or time-to-first-spike coding.
This setup is motivated by three considerations: \textbf{(i)} It sets the focus on purely temporal spike coding, which has increasingly been used for machine learning applications \cite{goltz2021fast,stanojevic2023exact,goltz2024delgrad,lunghi2024investigation}. \textbf{(ii)} The nLIF model is analytically tractable, enabling provable theoretical results. \textbf{(iii)} As a special case of the widely used LIF model (see \cref{si:LIF}), it provides a principled starting point for extending our results in future work. 
We demonstrate this by providing simulations for single-spike LIF neurons with finite membrane time constant ($\taum = 2 \taus$).
In general, causal pieces are computed similarly for standard LIF and other more complex neuron models.

The definition of a feedforward nLIF neural network is given in \cref{def:nLIF}.
\begin{challengebox}
\begin{definition}[nLIF]\label{def:nLIF}
    Let $L \in \mathbb{N}$, $\ell \in [1,L]$, $N_\ell \in \mathbb{N}$ be the number of neurons per layer $\ell$, $\taus \in \mathbb{R}^+$ be the synaptic time constant, $\thresh \in \mathbb{R}$ be the threshold, and
    $t^{(0)} \in \mathbb{R}^{N_0}$, $N_0 \in \mathbb{N}$, be the inputs to the neural network. For $i \in [1, N_\ell]$, $j \in [1, N_{\ell-1}]$, let $W^{(\ell)}_{ij} \in \mathbb{R}$ be the synaptic weights from layer $\ell-1$ to $\ell$.
    % , and $t^{(\ell)}_j \in \mathbb{R} \cup \{\infty\}$ the spike time of neuron $j$ in layer $l$.
    Then the membrane potential $u^{(\ell)}_i \in \mathbb{R}$ of a neuron $i$ in layer $\ell$ at time $t \in \mathbb{R}$ is given by:
\begin{equation}\label{eq:u}
    u^{(\ell)}_{i}(t) = \sum_{t^{(\ell-1)}_j \leq t} W^{(\ell)}_{ij} \left[ 1 - \exp \left(-\frac{t-t^{(\ell-1)}_j}{\taus} \right) \right]  \,.
\end{equation}
The spike time $t^{(\ell)}_i$ of a neuron $i$ in layer $\ell$ is defined as $t^{(\ell)}_i = \mathrm{inf}\{t: u^{(\ell)}_i(t) = \thresh\}$.
\end{definition}
\end{challengebox}

The output spike time of an nLIF neuron (and LIF with $\taum = 2 \taus$) can be calculated analytically by finding, given a set of input spike times and weights, the corresponding \textit{causal set} (see \cref{si:analytical_spike}) \cite{goltz2021fast}.
The causal set contains the indices of all pre-synaptic neurons that cause the output spike time, i.e., it is the sorted set of neurons whose input spikes occur before the output spike (\cref{def:cset}).
\vspace{-4.pt}
\begin{challengebox}
\begin{definition}[Causal set]\label{def:cset}
    Let $t^{(\ell)}_i \in \mathbb{R} \cup \{\infty\}$ be the spike time of a neuron receiving $N_{\ell-1} \in \mathbb{N}$ input spikes at times $t^{(\ell-1)}_j$ for $j \in [1, N_{\ell-1}]$.
    Then the corresponding causal set is given by $\causal^{(\ell)}_i(t^{(\ell-1)}_1, ..., t^{(\ell-1)}_{N_{\ell-1}}) = \left(j_1, ..., j_k\right)$ with
 $\left\{j_1, ..., j_k\right\} = \{j: t^{(\ell-1)}_j < t^{(\ell)}_i \}$\ and $j_1 \leq ... \leq j_k$ if\ $t^{(\ell)}_i < \infty$ and $\causal^{(\ell)}_i(t^{(\ell-1)}_1, ..., t^{(\ell-1)}_{N_{\ell-1}}) = ()$ otherwise.
\end{definition}
\end{challengebox}
\vspace{-4.pt}
Causal sets implicitly depend on the neural network weights through $t_i^{(\ell)}$.
For deep feedforward SNNs, the concept is generalised by introducing \textit{causal subnetworks} (\cref{def:cpath}).
\vspace{-4.pt}
\begin{challengebox}
\begin{definition}[Causal subnetwork] \label{def:cpath}
Let $L \in \mathbb{N}$, $\ell \in [1,L]$, $N_\ell \in \mathbb{N}$, $N_0 \in \mathbb{N}$. Further, let $\causal_i^{(m)}$ be the causal set of neuron $i \in [1, N_m]$ in layer $m \in [1, L]$. Then for a subset $I \subseteq [1, N_\ell]$ of neurons in layer $\ell$, the causal subnetwork  $\paths^{(\ell)}_{I}(t^{(0)})$ given inputs $t^{(0)} \in \mathbb{R}^{N_0}$ is defined recursively:
\begin{equation}
    \paths^{(\ell)}_{I,n-1} = \left(\causal^{(n-1)}_j: j \in \causal \text{\ for \ } \causal \in \paths^{(\ell)}_{I,n}\right)\,,
\end{equation}
with $\paths^{(\ell)}_{I,\ell} = \left(\causal^{(\ell)}_i: i \in I \right) \ \ \ \ \text{and} \ \ \ \ n \in [1, \ell]$.
\end{definition} 
\end{challengebox}
\vspace{-4pt}
As depicted in \cref{fig:Intro}A, a causal subnetwork refers to the subset of neurons and connections that influenced the output spike times of neurons $i \in I$ of layer $\ell$, given inputs $t^{(0)}$.
In this work, we represent it as a list of lists: for each layer, we include a list containing the causal sets of neurons in this layer that contributed to the spike times of neurons in $I$. 
It can be calculated from the observed spike times and connectome alone (see the algorithm in \cref{alg:pieces}), but depends implicitly on the weights.
An illustration of causal sets and causal subnetworks can be found in \cref{si:sets_illu}.

Using these definitions, we can now introduce the concept of causal pieces, which we later demonstrate to be a useful tool for analysing the computational properties of SNNs.
For a subset $I$ of neurons in layer $\ell$ of a feedforward SNN, the causal piece is a region in the joint input and parameter space for which the causal subnetwork remains fixed, meaning that the spike times of neurons $i \in I$ depend on the same subnetwork (\cref{fig:Intro}A) within this region. 
Formally, using \cref{def:cset,def:cpath} we define a causal piece as follows (\cref{def:cpiece}):
\vspace{-4.pt}
\begin{challengebox}
\begin{definition}[Causal piece]\label{def:cpiece} Let $L \in \mathbb{N}$, $\ell \in [1,L]$, $N_\ell \in \mathbb{N}$, $t_0 \in \mathbb{R}^{N_0}$ be the input spike times to the SNN with $N_0 \in \mathbb{N}$, and weights $W \in \mathbb{W} =  \mathbb{R}^{N_0 \cdot N_1} \times ... \times \mathbb{R}^{N_{L-1} \cdot N_L}$. Then for a subset $I \subseteq [1, N_\ell]$ of neurons from layer $\ell \in [1, L]$, we call $\mathbb{P}[\paths^{(\ell)}_I]$
the causal piece associated to $\paths^{(\ell)}_I$:
\begin{center}
$
\mathbb{P}[\paths^{(\ell)}_I] = \{( t_0, W) \in \mathbb{R}^{N_0} \times \mathbb{W} \colon$ given $t_0$ \& $W$, the neurons $i \in I$ have causal subnetwork $\paths^{(\ell)}_I\}
$
\end{center}
\label{def:4}
\end{definition}
\end{challengebox}
\vspace{-4pt}
Throughout this paper, we often consider causal pieces for networks with fixed weights. 
In such cases, the causal piece is only defined by the inputs and reduces to $\mathbb{P}[\paths^{(\ell)}_I] \subseteq \mathbb{R}^{N_0}$. 

Under the perspective of \cref{def:4}, the output spike times of an SNN are piecewise continuous, non-linear functions (\cref{fig:Intro}B) of the inputs, with each region corresponding to a distinct causal piece.
For nLIF networks, within each piece, the output spike times are Lipschitz continuous with respect to both the input spike times and weights -- a useful property for gradient-based training.
When transitioning between causal pieces -- for example, by varying the input to an nLIF neuron -- the output spike time may change continuously, discontinuously, or become undefined (i.e., the neuron ceases to spike), depending on how the causal sets of neurons change (see \cref{si:theoremLipschitz} for details).

\section{Results}
In the following, we first prove that the number of causal pieces provides a lower bound for the approximation error of a feedforward nLIF SNN.
We then continue by demonstrating how to count them.
The theoretical results are complemented by simulations, showing, in particular, that a high number of causal pieces on the training samples at initialisation correlates with training success (\cref{fig:PiecesVsAcc}).

\subsection{The number of causal pieces is a measure of expressivity}

The approximation error of an nLIF network, i.e., how well a given function can be approximated, is lower bounded by an expression depending inversely on the number of causal pieces -- meaning that more causal pieces result in potentially more expressive SNNs (\cref{theorem:nLifbound}; for a proof, see \cref{si:proofBound}).
\cref{theorem:nLifbound} proves a lower bound on the approximation error on an interval $[a,b]$. 
For a similar lower bound on a high-dimensional domain, it is enough to exhibit one subset of the domain on which the error is already large, since for every line $s \subset [a,b]^d$ it holds that 
 $\|\Phi - g \|_{L^\infty\left([a,b]^d\right)} \geq \|\Phi - g \|_{L^\infty\left(s\right)}$.
Most importantly, this result is valid even if $\Phi$ has any discontinuous behaviour.
\vspace{-3.pt}
\begin{challengebox}
\begin{theorem}[Approximation bound] \label{theorem:nLifbound}
Let $-\infty < a < b <\infty$, $g \in C^3([a,b])$ so that $\exp{(g/\taus)}$ is not affine. Then there exists a constant $c > 0$ that only depends on $\taus \int_a^b \sqrt{|\frac{\mathrm{d}^2}{\mathrm{d}x^2}e^{g(x) /\ \taus}|} \mathrm{d}x$ and a constant $\zeta > 0$ only depending on the maximum of $\mathrm{max}_x\left(e^{\Phi(x) /\ \taus}\right)$ and $\mathrm{max}_x\left(e^{g(x) /\ \taus}\right)$ so that
\begin{equation}
    \|\Phi - g \|_{L^\infty\left([a,b]\right)} > \frac{c}{\zeta} p^{-2} 
\end{equation}

for all feedforward nLIF neural networks $\Phi$ (\cref{def:nLIF}) with $p$ causal pieces and synaptic time constant $\taus$. 
\end{theorem}
\end{challengebox}
\vspace{-3pt}

This bound is also valid for nLIF neurons that spike multiple times (with a simple reset mechanism) due to a recently proven transference principle \cite{dold2026equivalence}.
Through this principle, an implied notion of causal pieces can be defined for multi-spike SNNs by constructing equivalent single-spike SNNs.

\subsection{Estimating the total number of causal pieces}\label{sec:perceptron}

Since the number of causal pieces is a measure of the expressivity of nLIF neural networks, it is of substantial interest to estimate this number.
As every causal piece is characterised by a unique causal subnetwork, one way is to calculate the total number of causal subnetworks that can be formed.
For a single nLIF neuron with $N$ total inputs, a naive upper bound for the number of causal pieces is therefore $2^{N}-1$ (the number of non-empty subsets that can be formed from a set of $N$ elements).

\begin{figure*}[t!]
    \centering
    \includegraphics[width=1\textwidth]{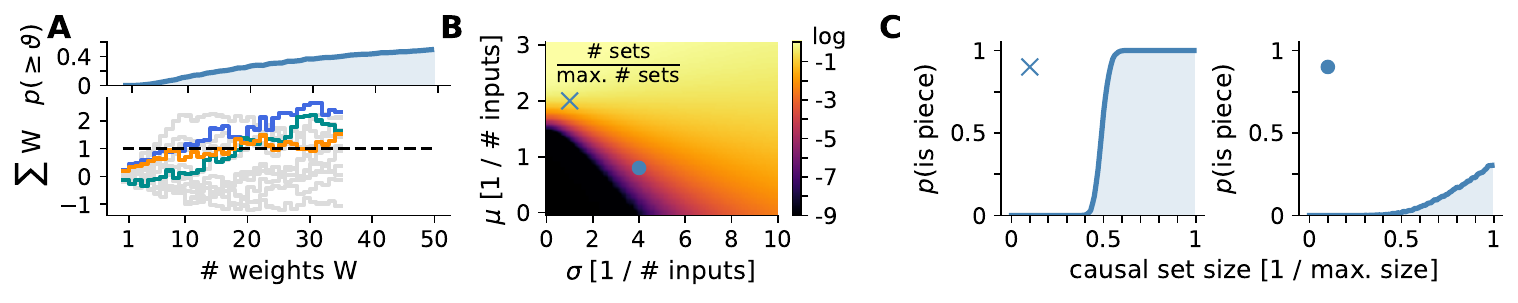}
    % \vspace{-5mm}
	\caption{Estimating the number of causal pieces.
    \textbf{(A)} The probabilities $p^q_k$ are obtained by counting how many trajectories (cumulative sum of weights) are above the threshold at step $k$. The top panel shows the corresponding values of $p^q_k$, where $k$ is the number of weights.
    \textbf{(B)} Estimated number of pieces for weights sampled from normal distribution with different mean (y-axis) and standard deviation (x-axis). Colours are shown in log-scale.
    \textbf{(C)} $p^q_k$ for two points in (B), denoted by markers.}
	\label{fig:RW}
\end{figure*}

However, not all of these subsets will be valid causal sets. In case of nLIF neurons, this means that the sum of the respective input weights might not exceed the threshold.
We obtain an improved upper bound by calculating the probability that, given weights sampled from a static random distribution $q$, the sum of $k$ weights exceeds the threshold, denoted by $p^q_k$.
This is equivalent to the probability of a discrete random walk with continuous random step sizes (i.e. the weights) being above the threshold at step $k$ (\cref{fig:RW}A).
This criterion is sufficient, as we can freely choose the inputs: all inputs of neurons in the causal set spike at the same time, while neurons not part of the set spike after the output neuron (\cref{si:randomWalk}).
The number of causal pieces $\eta^q$ is then upper bounded by $\eta^{q} \leq \sum_{k=1}^{N} \binom{N}{k} p^{q}_k$.

We show the improved upper bound of the number of sets as a fraction of $2^N - 1$ in \cref{fig:RW}B for weights randomly initialised from Gaussian distributions with different mean and variance (using a Monte Carlo approach, see \cref{alg:mc} in the appendix).
For illustration purposes, $p^q_k$ is shown for two different $q$ in \cref{fig:RW}C.
The obtained results highlight two points: \textbf{(i)} the highest number of causal pieces is reached only for distributions with non-zero mean -- which is quite remarkable given that initialisation schemes in the literature, often borrowed from traditional deep learning, sample the weights from distributions with zero mean \cite{rossbroich2022fluctuation,bellec2018long,zenke2021remarkable,lee2016training,ding2022accelerating,che2024ettfs}.
However, for extreme distributions, e.g., with very high mean, the estimated number of pieces is only achieved for data distributions that are very different from those found in practice, which will be discussed more thoroughly in the next section.
\textbf{(ii)} With increasing variance, results tend to improve even if the mean is set non-optimally. 
In fact, one can show that in the limit of large variance, the number of pieces is lower bounded by an expression proportional to $N^{-3 / 2}$ (\cref{theorem:lbound}):
\vspace{-4.pt}
\begin{challengebox}
\begin{theorem}[Number of pieces in limit]\label{theorem:lbound}
    Let $q$ be a symmetric and continuous probability distribution with mean $\mu < \infty$ and variance $\sigma^2$, and $W_j \sim q$ for $0 \leq j < N$. 
    In the limit $\frac{\mu}{\sigma} \rightarrow 0$ and $\frac{\thresh}{\sigma} \rightarrow 0$, the number of causal pieces is lower bounded by
    \begin{equation}
        \eta^q \geq \frac{2^N-1}{2 N \sqrt{\pi \cdot (N - \frac{2}{3})}} \,,
    \end{equation} 
\end{theorem}
\end{challengebox}
\vspace{-4pt}
which is, quite remarkably, valid for all such probability distributions.
This is a direct consequence of the Sparre Andersen theorem for random walks \cite{andersen1954fluctuations,majumdar2010universal}, see \cref{si:proofSparre}.
While this result is only valid when all weights follow the same distribution, this can be guaranteed at initialisation, explaining the observed growth in the number of causal pieces with increased variance (\cref{fig:RW}B).

For deep SNNs, the number of causal pieces corresponds to the number of distinct routes through which spikes can propagate from the inputs to the outputs (\cref{def:cpath}).
For nLIF networks with $\{N_1, ..., N_\ell, 1\}$ neurons per layer, we find in \cref{si:piecesComb} that a naive upper bound for the number of pieces of the output neuron is $\eta^q \leq 2^{\prod_{i=1}^\ell N_i} \leq 2^{N^\ell}$, where $N = \mathrm{max}\{N_1, ..., N_\ell, 1\}$.
This differs from ReLU ANNs, for which linear pieces are determined by neuron activation patterns rather than routes, with known upper bounds for the number of pieces scaling exponentially in depth \cite{montufar2014number,telgarsky2016benefits} (or the total number of neurons \cite{hanin2019complexity}).
However, it remains to be seen whether SNNs with such a large number of pieces can be constructed, although \cref{fig:Intro}C suggests quite dramatic increases in the number of causal pieces by adding even a single hidden layer.
Since for general LIF neurons, weights also have to sum above threshold, similar estimates for the expected number of pieces apply.

\subsection{The practically relevant number of causal pieces}

In practice, even for single neurons we expect the number of pieces to be below the improved bound we found, as most of these pieces will not be traversed when given realistic input data (i.e., not all inputs being identical).
Moreover, the total number of pieces may be irrelevant for the learning problem at hand if a large fraction of the pieces occupy parts of the domain that are not populated by data. For example, \cref{fig:Intro}C shows that the density of pieces can change dramatically throughout the domain.
Thus, we propose an alternative approach to counting causal pieces which is more aligned to practical scenarios and less resource demanding: given a dataset, we count only the number of pieces that contain at least one data point. 
In the following, we only use this approach to count the number of causal pieces.
An algorithm for counting causal pieces is provided in \cref{alg:pieces}.
Using this approach, the complexity of calculating causal pieces scales linearly with the number of neural network parameters and training samples (see \cref{si:time} for simulation results on compute time for a single data sample) -- although this can be optimised by parallelising the calculation per layer.

\subsection{The number of practically relevant causal pieces correlates with training success}

\begin{figure*}[b!]
    \centering
    \includegraphics[width=1\textwidth]{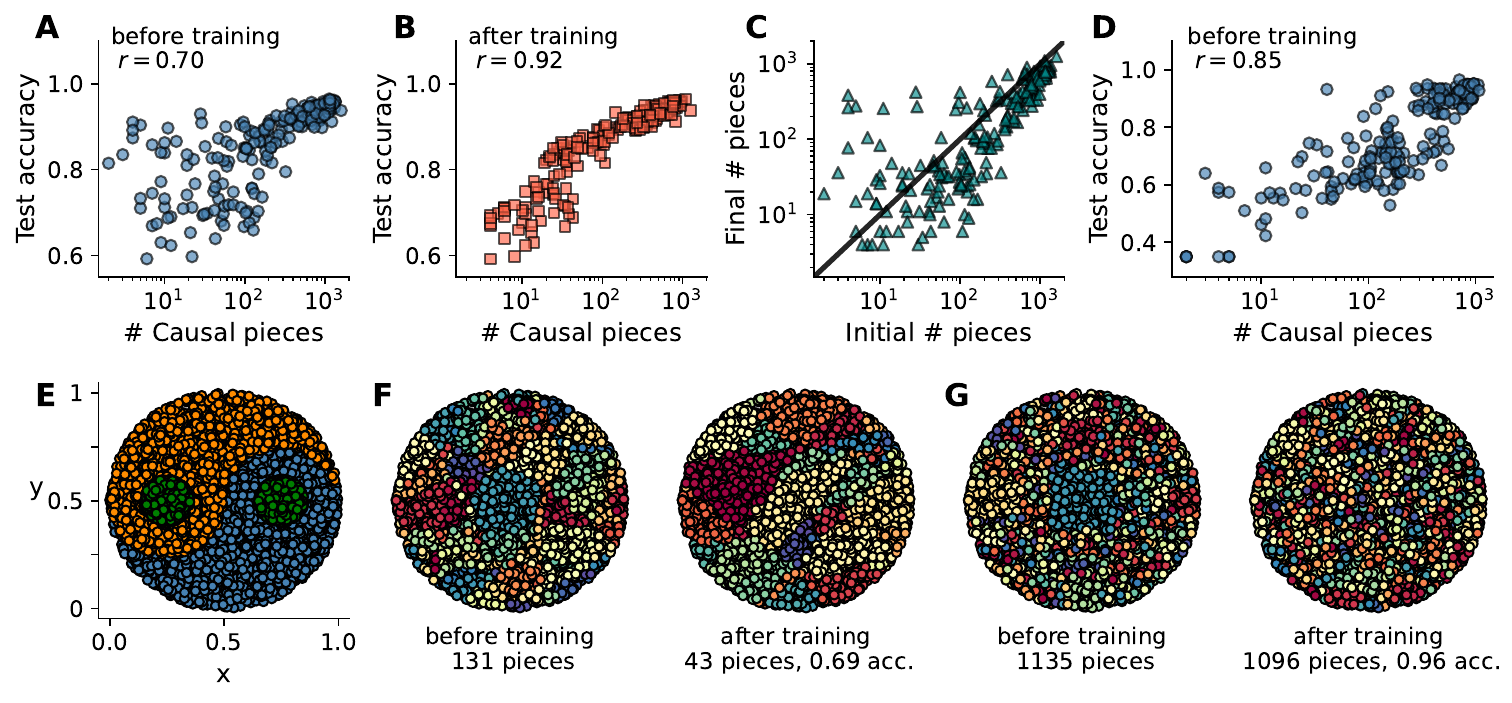}
	\caption{Network initialisation strongly affects training success. 
    \textbf{(A)} The logarithm of the number of pieces (here: of the output layer) at network initialisation strongly correlates with performance after training ($r = 0.70$, log-linear). Shown for $\taus = 0.5$.
    \textbf{(B)} Same as (A), but with the number of pieces after training. 
    \textbf{(C)} Number of pieces before and after training. The diagonal indicates no change in pieces.
    \textbf{(D)} Same as (A), but for nLIF neurons with time constant $\taus = 0.1$.
    \textbf{(E)} Illustration of the Yin-Yang dataset with three classes: the two halves and the dots.
    \textbf{(F)} Causal pieces (each piece is indicated by a different colour) of the hidden layer for a poor initialisation, evaluated using only training samples.
    \textbf{(G)} Same as (F), but for one of the best initialisations.}	
	\label{fig:PiecesVsAcc}
\end{figure*}
\begin{figure*}[b!]
    \centering
    \includegraphics[width=1\textwidth]{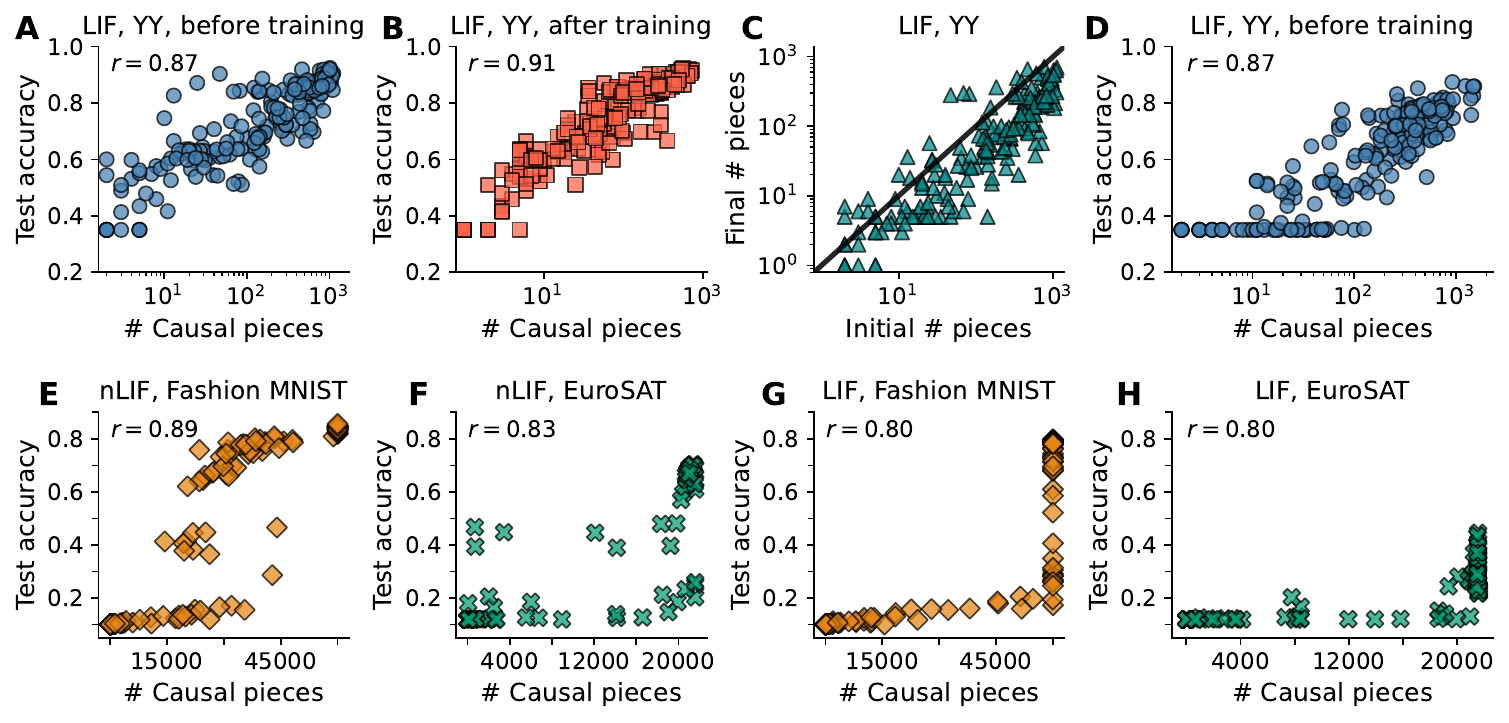}
	\caption{Experimental validation using more complex neuron models and datasets. 
    \textbf{(A-C)} Same as \cref{fig:PiecesVsAcc}A-C, but for LIF neurons with $\taum = 0.5$ and $\taus = 0.25$.
    \textbf{(D)} Same as \cref{fig:PiecesVsAcc}D, but for LIF neurons with $\taum = 0.1$ and $\taus = 0.05$. 
    \textbf{(E-F)} The number of causal pieces at initialisation correlates with final test accuracy for nLIF SNNs trained on Fashion MNIST and EuroSAT. 
    \textbf{(G-H)} Same as E-F, but for LIF SNNs with $\taum = 0.5$ and $\taus = 0.25$. 
    E-H are plotted on linear axes for easier readability. Reported correlations are also calculated using the untransformed number of causal pieces. Log-linear correlations are (from E to H): 0.74, 0.69, 0.62, and 0.62.}
	\label{fig:fashion}
\end{figure*}

The initialisation scheme of parameters is crucial for training both ANNs and SNNs. 
Although for SNNs, schemes derived experimentally or adopted from ANNs have been successfully applied, a recent study highlighted the lack of a principled approach for identifying initialisation schemes that facilitate the training of SNNs \cite{rossbroich2022fluctuation}.
As a first application, we demonstrate that the number of causal pieces at initialisation, evaluated only using training samples, is a strong predictor of training success.
This is motivated by \cref{theorem:nLifbound}, which provides a lower bound of the approximation error, meaning that no nLIF SNN with 
$p$ pieces can reach an error lower than this bound. Thus, when an SNN has only few pieces at initialisation, this bound is high, and lower errors are initially not reachable. In contrast, if the number of pieces is high at initialisation, this particular approximation-theoretic obstruction is reduced; whether optimization actually reaches a low-error solution remains dependent on the training dynamics.

We trained 200 shallow nLIF networks with $[4, 30, 3]$ neurons, where the last layer is a linear readout layer (``affine decoder'' \cite{neuman2024stable}) that ignores neurons that do not spike, on the Yin-Yang dataset \cite{kriener2022yin} (with 5000 training samples). 
Training was performed using exact error backpropagation on analytically calculated spike times (via standard autograd in pyTorch), as introduced in \cite{mostafa2017supervised,goltz2021fast} -- although we do not use any weight regularisation like ``weight-bumping'' \cite{mostafa2017supervised,goltz2021fast,goltz2024delgrad}.
Inputs were presented to the network as single spike times.
To guarantee networks with a large variety of causal pieces after initialisation, we sampled weights from a normal distribution with randomly sampled mean and variance (see \cref{si:details}).
As shown in \cref{fig:PiecesVsAcc}A,B, both the number of causal pieces of the last layer before and after training (evaluated using only training samples) strongly correlate with the final accuracy achieved on the test split.
Before training, all SNNs perform around chance level independent of the initial number of causal pieces.

Interestingly, we find that it seems hard to recover from a poor initialisation with a low number of pieces through training (\cref{fig:PiecesVsAcc}C).
Although some SNNs with an initially small piece count achieve a higher -- but not the best -- test accuracy, this is only possible because the number of pieces was significantly increased during training, in agreement with \cref{theorem:nLifbound}.
However, for most SNNs, the piece count decreases during training.
A similar result is obtained for smaller time constants (\cref{fig:PiecesVsAcc}D), where SNNs with an initial low piece count perform even worse on average.
Due to the decreased time constant, inputs can drive the neuron's membrane potential faster above the threshold, meaning that weights have to be carefully tuned to allow integration of late input spikes -- thus complicating the creation of new causal pieces during training.
Examples of the causal piece structure on the training data of the Yin-Yang dataset are shown for the hidden layer of an SNN achieving low (\cref{fig:PiecesVsAcc}F) and high performance (\cref{fig:PiecesVsAcc}G) -- clearly highlighting the difference in the number of causal pieces before and after training (separate clusters with similar colours are different causal pieces).
We also find that initially, training samples frequently switch between different causal pieces -- a reorganisation that slows down and stabilises in later epochs.
For SNNs with a low initial piece count, reorganisation eventually dies out, while for SNNs with high initial count, it never stops (see \cref{si:reorg}).

A similar trend is observed for single-spike LIF neurons with finite membrane and synapse time constant (\cref{fig:fashion}A-D).
In this case, recovery from an initial low piece count is even rarer, most likely because creating causal pieces is harder than in the nLIF case: a valid causal set not only requires that the weights of input spikes exceed the threshold, but that the input spikes also co-occur sufficiently close together in time.
This illustrates that, while \cref{theorem:nLifbound} is restricted to nLIF neurons, the applicability of causal pieces extends to more complex neuron models with leaky dynamics.

These results are not restricted to the Yin-Yang dataset.
In \cref{fig:fashion}E-H, we observe the same trend for shallow SNNs with 200 hidden neurons (and linear readout) trained via exact gradient descent on the Fashion-MNIST dataset \cite{xiao2017fashion} and the EuroSAT RGB benchmark \cite{helber2019eurosat}, a land cover classification task using real images from the European Space Agency's (ESA) Sentinel-2 satellite.
Consequently, we hypothesise that the number of causal pieces is a promising metric for identifying good initialisation schemes for SNNs -- although it is certainly not the sole predictor of training success.
For image data, pixel values were translated into spike time latencies before feeding them into the network (see \cref{si:details}).
\begin{figure*}[b!]
    \centering
    \includegraphics[width=1.\textwidth]{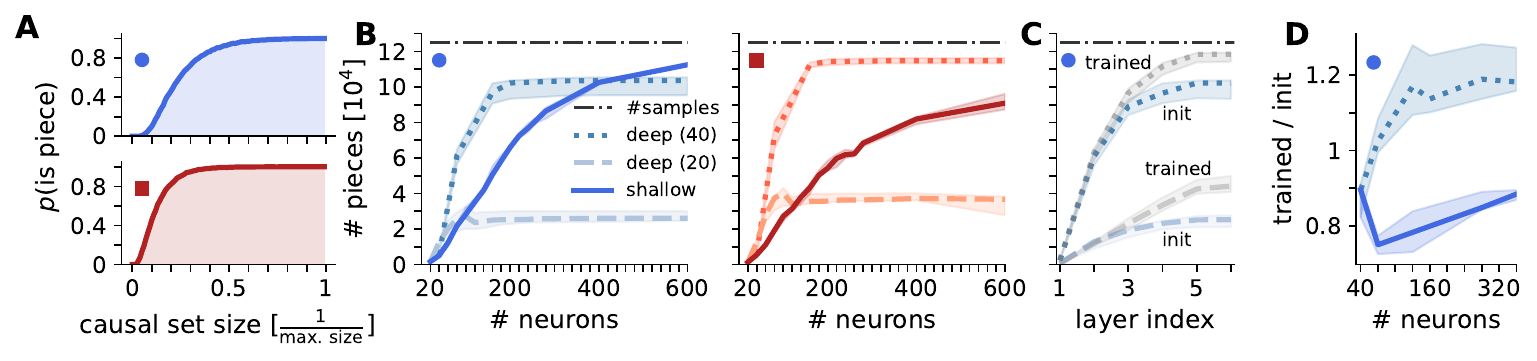}
    % \vspace{-5mm}
	\caption{Width-and depth dependence of causal pieces.
    \textbf{(A)} $p^q_k$ of the optimised (top, dot) normal, and (bottom, square) uniform initialisation.
    \textbf{(B)} Number of pieces for shallow and deep networks. The maximum number, which is the number of input samples used to evaluate the number of causal pieces, is shown as a dash-dotted line.
    \textbf{(C)} Number of pieces per layer in a single network, before and after training.
    \textbf{(D)} Increase in the total number of pieces for deep and shallow networks. 
    Markers denote results that belong together. We show medians (lines) and quartiles (shaded areas).}	
	\label{fig:Deep}
\end{figure*}

\subsection{Increasing the number of pieces by going wide or deep}

As seen in the previous subsection, a large initial number of pieces is strongly associated with successful training in the settings studied here.
Therefore, it is a natural question to ask how this number can be increased. 
From the previous results, an obvious option is to optimise the weight initialisation to yield networks with many pieces.
We investigate this for networks ($[4, 100, 3]$ neurons) with weights initialised randomly from either a Gaussian or a uniform distribution, using a Yin-Yang dataset obtained from a $400 \times 400$ grid on the data domain.
We chose a larger dataset here to properly probe the number of causal pieces.
In case of a Gaussian distribution, the weights projecting into layer $\ell \in \mathbb{N}$, $W^{(\ell)} \in \mathbb{R}^{n_\ell\times n_{\ell-1}}$, are initialised by sampling from $\mathcal{N}(\alpha_0 \cdot n_{\ell-1}^{-\alpha_1}, \left[\alpha_2 \cdot n_{\ell-1}^{-\alpha_3}\right]^2)$ where $n_{\ell}$ is the number of neurons in layer $\ell$.
Similarly, in case of a uniform distribution, weights are sampled from $\mathcal{U}(-v_0 + v_1, v_0 + v_1)$ with $v_0 = \beta_0 \cdot n_{\ell-1}^{-\beta_1}$ and $v_1 =  \beta_2 \cdot n_{\ell-1}^{-\beta_3}$.
The parameters $\alpha_i$ and $\beta_i$ ($i \in [0,3]$) are found using a simple evolutionary algorithm that maximises the number of causal pieces (\cref{si:evo}).
For this specific setup, we found $\alpha_0 = 1.69$, $\alpha_1 = 0.79$, $\alpha_2 = 1.13$, $\alpha_3 = 0.49$ and $\beta_0 = 1.85$, $\beta_1 = 0.39$, $\beta_2 = 1.02$, $\beta_3 = 0.54$.
The corresponding probabilities $p^q_k$ of these weight initialisations are shown in \cref{fig:Deep}A.
As for the single neuron case, the weight distributions feature non-zero means.
We visualise the causal pieces for a single output neuron in \cref{si:piecestructures}.

Another option to adjust the number of pieces is to change the width and depth of the SNN, as shown in \cref{fig:Deep}B,C. 
We present three scenarios: (line) a shallow network where the width is steadily increased by increments of 20 neurons, (dashed) a deep network, where in each increment an additional hidden layer with 20 neurons is added, and (dotted) the same as for dashed, but with 40 neurons per hidden layer.
Results are shown for the two distributions found using evolutionary optimisation.
For the shallow network, the number of pieces grows consistently with increased network width, although slower than for deep networks and with saturation setting in for very wide networks. 
In case of deep networks, the number of pieces grows rapidly initially, but then stagnates to a constant number of causal pieces.
The effect is more pronounced if the hidden layers are wider, with a much stronger increase and final number of causal pieces for the network with 40 neurons per layer.
Different from the expected exponential increase, we rather see a logistic growth.
In fact, fitting logistic curves of the form ${\gamma_0}/{(\gamma_1 + e^{-\gamma_2 N})}$ with $\gamma_i \in \mathbb{R}$ and $N$ the number of neurons, we get a median relative error of $4\cdot10^{-2}$ (shallow), $2\cdot 10^{-2}$ (deep 20), and $2\cdot 10^{-2}$ (deep 40) for the Gaussian initialisation, and $9\cdot10^{-2}$ (shallow), $2\cdot10^{-2}$ (deep 20), and $5\cdot10^{-3}$ (deep 40) for the uniform one.
The saturation for (deep 20) most likely occurs due to a diminishing effect of pieces being split by consecutive layers, similar to how for ReLU ANNs, an exponential growth in linear pieces with depth is often not observed in practice \cite{hanin2019complexity}.
For all other cases, saturation occurs as we reach the maximum number of causal pieces that can be counted using the data samples.

In \cref{fig:Deep}C, we show the number of pieces per layer for a network with 5 hidden layers. 
Similarly to how initially adding hidden layers increased the number of pieces drastically in \cref{fig:Deep}B, the highest increase is seen in the first few layers, with diminishing returns in deeper layers. 
In contrast, if we compare the number of pieces per layer before and after training, we find a slight increase in the number of causal pieces for deep layers.
If we just focus on the total number of pieces of the whole network, we find that shallow networks end up with less pieces than at initialisation, while deep networks end up with more (\cref{fig:Deep}D).
Most likely, this is because in a deep network, the number of pieces can be optimised by improving the misalignment of pieces between consecutive layers.

\subsection{Large number of causal pieces in spiking neural networks with only positive weights}

Inspired by \cite{neuman2024stable}, we now study the case of SNNs with only excitatory neurons.
In the mammalian neocortex, around 80\% \cite{nieuwenhuys1994neocortex} of neurons are excitatory, i.e., their synapses only excite other neurons, which is equivalent to neurons having only positive outgoing weights in our nLIF neural networks.
Although having only positive weights seems limiting at first, it comes with a significant advantage: controlling for continuity between causal pieces becomes much easier.
In fact, the network is globally Lipschitz continuous as long as for each neuron, the input weights have a sum larger than the threshold -- which can be easily enforced during training, e.g., through a regularisation term.
The global Lipschitz constant of a neural network can be used to derive its covering number, which provides an upper bound for the network's generalisation error \cite{petersen2024mathematical,neuman2024stable} (see \cref{theorem:lipschitz} in \cref{si:theoremLipschitz}).

\begin{figure*}[h!]
    \centering
    \includegraphics[width=1.\textwidth]{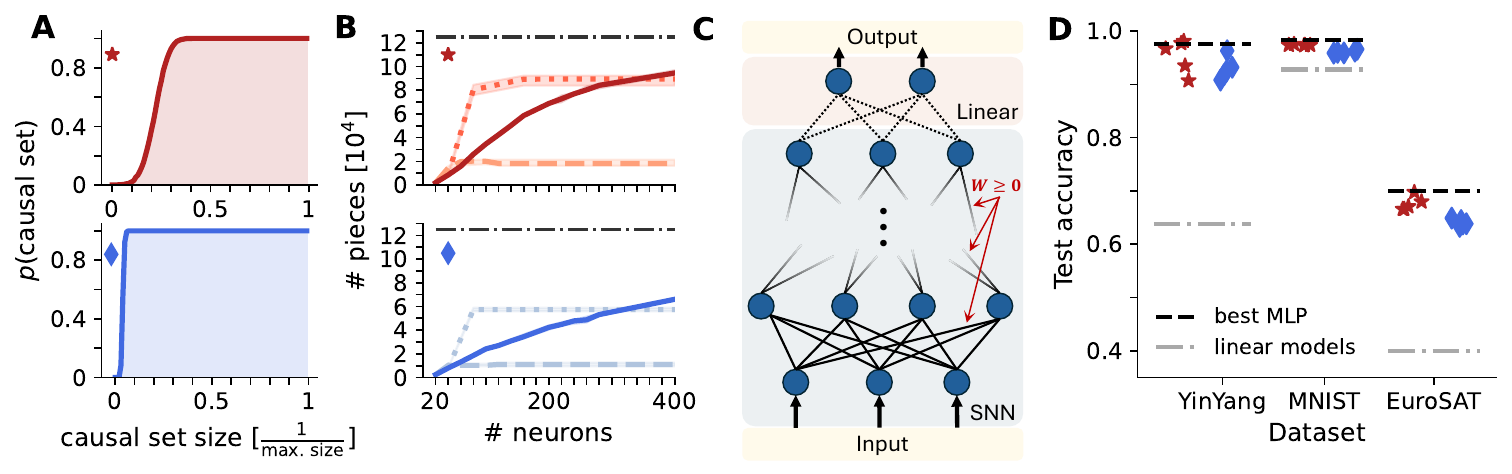}
    % \vspace{-5mm}
	\caption{SNNs with only positive weights. 
    \textbf{(A)} $p^q_k$ for (top, star) lognormal, and (bottom, diamond) uniform initialisation.
    \textbf{(B)} Number of pieces for shallow and deep networks. Labels as in \cref{fig:Deep}B.
    \textbf{(C)} Used network architecture.
    \textbf{(D)} Performance on benchmarks. Each training run was repeated 5 times for different random seeds. Markers denote results that belong together.}
	\label{fig:PosSNN}
\end{figure*}

We again optimise the parameters of two initialisation distributions, this time a lognormal and a uniform distribution -- which both lead to networks with a similar number of pieces than for distributions with both positive and negative values.
Their respective $p^q_k$ probabilities are shown in \cref{fig:PosSNN}A.
Using these initialisation schemes, we train deep networks (two hidden layers with 200 and 100 neurons) composed of an SNN with positive weights and a single linear readout layer (with positive and negative weights, see \cref{fig:PosSNN}C) on three different benchmarks: Yin-Yang, MNIST, and EuroSAT RGB -- achieving performance levels similar to fully-connected ANNs, and far outcompeting linear models (\cref{fig:PosSNN}D).
An illustration of the causal pieces is shown in \cref{si:fig}.
\vspace{-4.5mm}

\section{Discussion}\label{sec:discussion}
The presented results demonstrate that causal pieces are not only a powerful tool for increasing our understanding of SNNs, but also for guiding the design of improved network architectures and training methods.
The causal piece framework naturally fits the discontinuous, event-based nature of SNNs. It enables a mathematically rigorous analysis of SNNs without requiring restrictive assumptions such as positive weights and highly abstract neuron models.
Across all reported experiments, we found that initialising weights from distributions with non-zero mean yields SNNs with the highest number of causal pieces -- a strategy also used in \cite{goltz2021fast} for single-spike LIF neurons. 

While previous work has briefly explored linear pieces in simplified spike-response models with linearly rising postsynaptic potentials \cite{singh2023expressivity}, our work lays the foundation for elevating this concept to more realistic neuron models.
The decomposition of the input and parameter space into causal pieces should, in principle, be applicable to any spiking neuron model commonly used in practice, especially since only the connectome and spike times of an SNN are required to obtain causal pieces.
Although our theory and experiments focused on single-spike coding, we note that key results -- such as the approximation bound in \cref{theorem:nLifbound} -- extend to the multi-spike setting through a transference principle \cite{dold2026equivalence}.
Moreover, this transference principle provides a promising path forward to define causal pieces for multi-spike SNNs and derive approximation bounds such as \cref{theorem:nLifbound} directly.
For spiking neurons with leak, the definition of causal sets, and hence causal subnetworks and pieces, remains unchanged (cf. \cite{goltz2021fast,comsa2020temporal}).
Accordingly, causal pieces can be readily evaluated for general LIF neurons with single-spike coding, which we demonstrated in simulations (\cref{fig:fashion}).
Similar to affine pieces, causal pieces are limited to feedforward neural networks (including convolutional networks).
The scope of our contribution is summarised in \cref{tab:claim_scope}.
\begin{table}[h!]
\centering
\caption{Summary of our work's scope.}
\label{tab:claim_scope}
\renewcommand{\arraystretch}{1.2}
\begin{tabular}{lp{0.68\linewidth}}
\toprule
\textbf{Defined} &
Causal sets, subnetworks, and pieces for feedforward, single-spike SNNs,
calculated from connectivity and observed spike times. The definitions are not
specific to nLIF dynamics and apply unchanged to standard LIF neurons. \\
\textbf{Proved} &
\cref{theorem:nLifbound}, local Lipschitz continuity within pieces (\cref{theorem:lipschitz}), and counting results for
feedforward nLIF networks. The limited multi-spike extension follows through the cited transference principle. \\
\textbf{Empirically supported} &
Correlation results for single-spike nLIF and finite-time-constant LIF networks
trained with exact spike-time gradients on the three reported static
classification benchmarks, including signed- and positive-weight neural networks. \\
\textbf{Not established here} &
Arbitrary multi-spike dynamics, surrogate-gradient training, and recurrent neural
networks. \\
\bottomrule
\end{tabular}
\end{table}
Extending mathematical results to more complex neuron models will require non-trivial adaptations of the proof strategy.
Generally, as long as the output spike time of a neuron can be rewritten as a linear function of its inputs (e.g., via a change of variables), the same proof strategy as presented in this work remains applicable -- which is true for other simple neuron models, such as the simplified spike-response model used in \cite{stanojevic2024high}.
Models with analytical solutions for spike times, such as LIF with $\taum = \taus$ or $\taum = 2\taus$ \cite{goltz2021fast}, provide a promising angle for generalising the proven theorems to neurons with leakage.
In any case, our experimental results with standard LIF neurons indicate that theorems such as \cref{theorem:nLifbound} may partially extend to more complex neuron models.

Another important property of neural networks is their ability to generalise to previously unseen data.
Causal pieces, like their counterparts linear pieces used for ANNs, are primarily a tool for assessing the approximation ability of neural networks though.
Typical approaches for bounding the generalisation error of neural networks use covering numbers \cite{neuman2024stable}, which can only be calculated for neural networks that are globally Lipschitz continuous -- a property that does not hold for SNNs in general.
However, novel measures based on causal pieces, e.g., comparing the number of pieces the training and validation data fall into, might provide novel insights into the generalisation capabilities of SNNs and ANNs alike.
As a first step, we already report test errors in simulations in this work.

To conclude, we are confident that this approach is a promising direction that has the potential to be applicable to a wide range of neuron models and enables principled comparisons across spiking neuron models as well as with ReLU-based ANNs.
Moreover, we believe that the usefulness of causal pieces extends beyond technical applications and domains, potentially providing novel ways to study biological neurons by analysing their causal piece structure derived from experimental data.

\begin{ack}
We would like to thank Julian Göltz for helpful feedback on the manuscript.
D.D. was funded by the Horizon Europe's Marie Skłodowska-Curie Actions (MSCA) Project 101103062 (BASE).
Simulations were performed on the Austrian Scientific Computing (ASC) cluster.
\end{ack}

\printbibliography
\addcontentsline{toc}{section}{References}
\medskip

%%%%%%%%%%%%%%%%%%%%%%%%%%%%%%%%%%%%%%%%%%%%%%%%%%%%%%%%%%%%
\newpage

\appendix

\section{Technical appendices and supplementary material}

\subsection{Additional figures and simulation results}\label{si:fig}

\subsubsection{Illustration: causal sets and causal subnetworks}\label{si:sets_illu}

\begin{figure*}[h!]
    \centering
    \includegraphics[width=0.85\columnwidth]{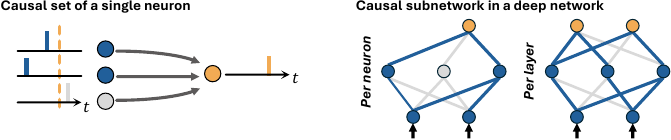}
	\caption{Left: A single output neuron (orange) receiving input from three neurons. Only those input neurons (blue) that spike before the output neuron (i.e., before the dotted line) are part of the causal set. Right: In deep networks, this corresponds to subnetworks (blue), here shown for a single output neuron (left), or the whole output layer (right).}
	\label{fig:SIsets}
\end{figure*}

\subsubsection{Time complexity}\label{si:time}

\begin{figure*}[h!]
    \centering
    \includegraphics[width=0.65\columnwidth]{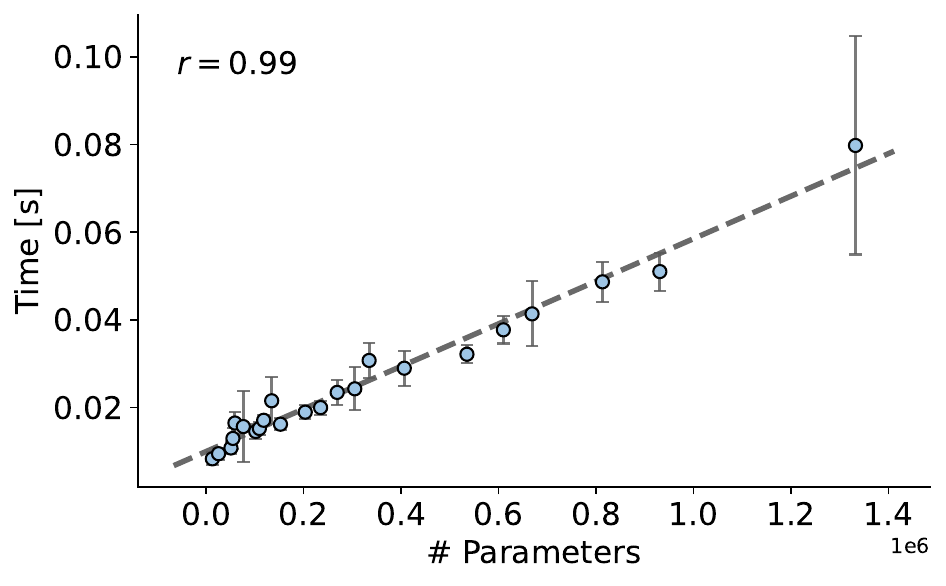}
	\caption{Wall-clock time required to calculate the causal piece for a single data sample, using MNIST. Network architectures range from a single to three hidden layers. For each network architecture, the average of 40 different runs is shown. Error bars show the standard deviation, and the dashed line is a linear fit.}
	\label{fig:SItc}
\end{figure*}

\subsubsection{Causal piece reorganisation during training}\label{si:reorg}

\begin{figure*}[h!]
    \centering
    \includegraphics[width=0.875\columnwidth]{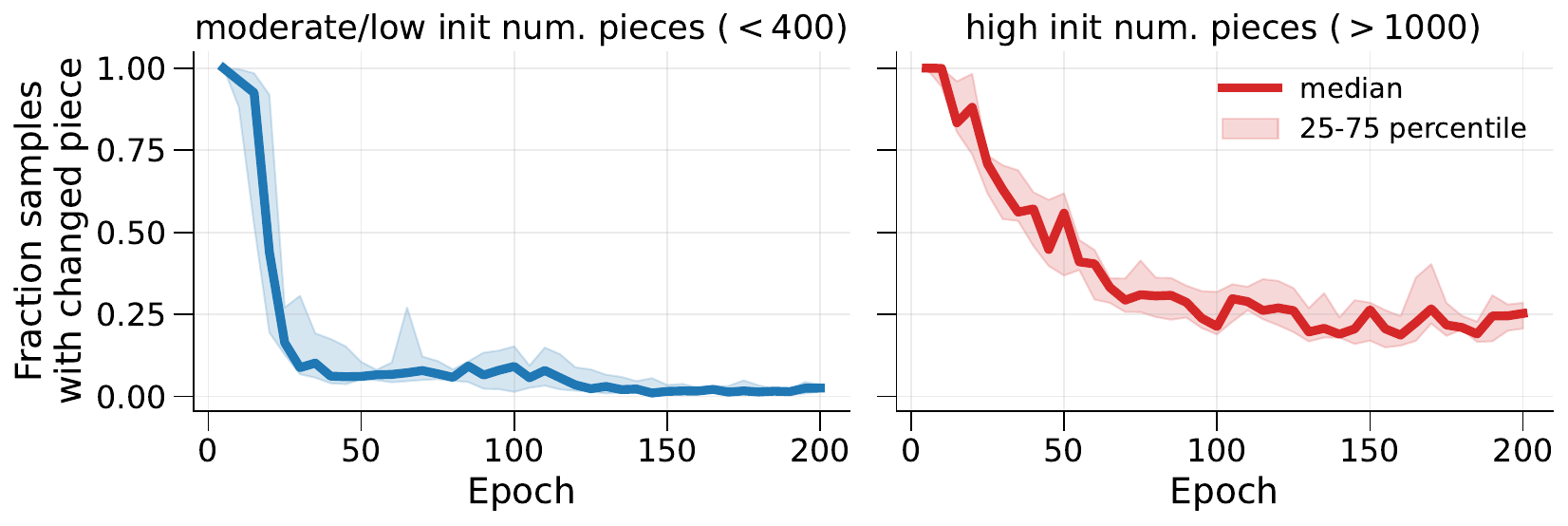}
	\caption{Fraction of data samples that switched causal piece compared to five epochs before. For $10$ nLIF SNNs trained on Yin-Yang with low/intermediate initial causal piece count, and $10$ nLIF SNNs with high initial count. The neurons use $\taus = 0.5$.}
	\label{fig:SIreorg}
\end{figure*}

\subsubsection{Ablation result: fraction of active neurons}

We test whether our observation can be explained via the average number of neurons that are active after initialisation, averaged over the training data samples.

Both for nLIF and LIF, the logarithm of the number of causal pieces at initialisation and the number of active neurons have a correlation coefficient of around 0.7-0.8, which is expected.
However, when training a linear model for predicting the chance-corrected test accuracy, we find that the logarithm of the number of causal pieces is more predictive than the fraction of active neurons (\cref{tab:piece_ablation}).

\begin{table}[h!]
\centering
\caption{Predictive power of feature sets for explaining final test accuracy. Lower Sum of Squared Residuals (SSR) is better.}
\label{tab:piece_ablation}
\begin{tabular}{llr}
\toprule
Model / Dataset & Feature set & SSR \\
\midrule
nLIF on Yin-Yang & Active fraction only & 5.17 \\
nLIF on Yin-Yang & Log initial piece count only & 2.95 \\
nLIF on Yin-Yang & Active fraction + log initial piece count & \textbf{2.63} \\
LIF on Yin-Yang & Active fraction only & 3.90 \\
LIF on Yin-Yang & Log initial piece count only & 1.17 \\
LIF on Yin-Yang & Active fraction + log initial piece count & \textbf{1.10} \\
\bottomrule
\end{tabular}
\end{table}

\subsubsection{Causal piece structures}\label{si:piecestructures}

\begin{figure*}[h!]
    \centering
    \includegraphics[width=0.95\columnwidth]{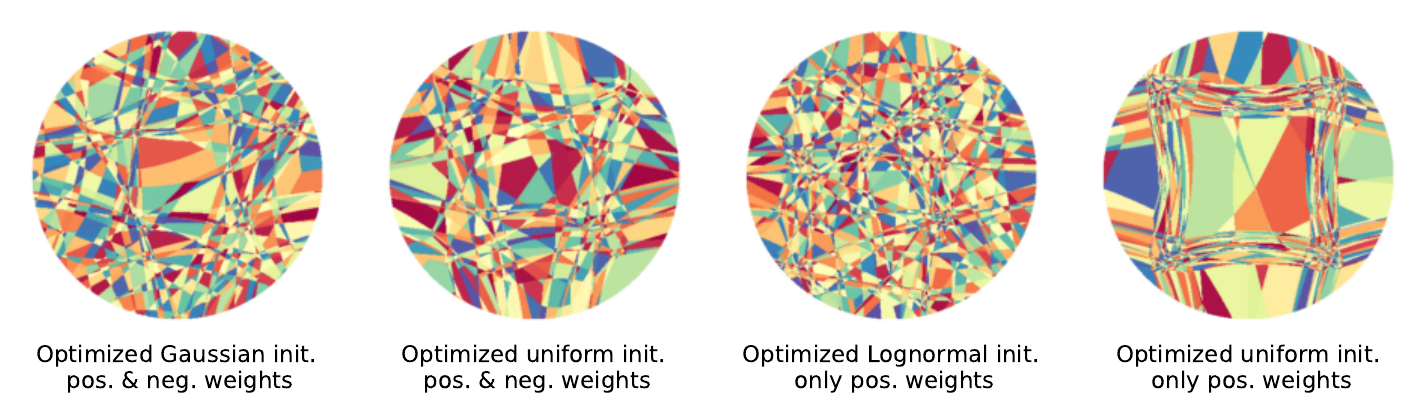}
	\caption{Causal pieces (coloured regions) of one of the output neurons for an nLIF neural network with $[4, 30, 3]$ neurons, using the initialisations obtained through evolutionary optimisation (\cref{fig:Deep} and \cref{fig:PosSNN}). Causal pieces are evaluated using a 400 $\times$ 400 grid on the data domain.}
	\label{fig:SI}
\end{figure*}

\subsection{Methods}

\subsubsection{Analytical solution of output spike time}\label{si:analytical_spike}

Given the causal set $\causal^{(\ell)}_i$, the corresponding output spike time $t^{(\ell)}_i$ of an nLIF neuron is \cite{mostafa2017supervised}
\begin{equation}\label{eq:spiketime}
        t^{(\ell)}_i = 
\begin{cases}
    \taus \ln \bigg(  \frac{\sum_{j \in  \causal^{(\ell)}_i} W_{ij}^{(\ell)} e^{t_j^{(\ell-1)} /\ \taus}}{\sum_{j \in  \causal^{(\ell)}_i} W_{ij}^{(\ell)} - \thresh}\bigg) \, & \text{if } \causal^{(\ell)}_i \neq \emptyset\,, \\
    \infty, & \text{else} \,,
\end{cases}
    \end{equation}
where the spike time is set to infinity if the input does not cause the neuron to spike.
To find the causal set in simulations, we use the iterative approach described in \cref{si:causalsets}.

\subsubsection{Relationship between nLIF and LIF neuron models}\label{si:LIF}

The current-based LIF neuron model with exponential synaptic kernel is given by
\begin{equation}\label{eq:dotuFull}
    \deriv{t} u^{(\ell)}_i(t) = \frac{1}{\taum} (u_\mathrm{rest} - u^{(\ell)}_i(t)) + \frac{1}{\taus}\sum_{j} W^{(\ell)}_{ij} \, \heaviside{t - t^{(\ell-1)}_j} \exp\left(-\frac{t-t^{(\ell -1)}_j}{\taus}\right) \,,
\end{equation}
where $u^{(\ell)}_i(t) \in \mathbb{R}$ is the membrane potential of neuron $i$ in layer $\ell$ at time $t \in \mathbb{R}$, $W^{(\ell)}_{ij} \in \mathbb{R}$ is the synaptic weight connecting neuron $j$ of layer $\ell - 1$ to neuron $i$ of layer $\ell$, $t^{(\ell - 1)}_j$ is the spike time of neuron $j$ in layer $\ell - 1$, $\taum \in \mathbb{R}^+$ and $\taus \in \mathbb{R}^+$ are the membrane and synaptic integration time constants, $\heaviside{\cdot}$ is the Heaviside function, and $u_\mathrm{rest} \in \mathbb{R}$ is the rest value of the membrane potential.

In the special case $\taum \gg \taus$, this simplifies to
\begin{equation}\label{eq:dotu}
    \deriv{t} u^{(\ell)}_i(t) = \frac{1}{\taus}\sum_{j} W^{(\ell)}_{ij} \, \heaviside{t - t^{(\ell-1)}_j} \exp\left(-\frac{t-t^{(\ell-1)}_j}{\taus}\right) \,,
\end{equation}
which can be solved for $u^{(\ell)}_i(t)$ by integration:
\begin{equation}\label{eq:u}
    u^{(\ell)}_{i}(t) = \integral{-\infty}{t}{\deriv{t'}u^{(\ell)}_i(t')}{t'} = \sum_{t^{(\ell-1)}_j \leq t} W^{(\ell)}_{ij} \left[ 1 - \exp \left(-\frac{t-t^{(\ell-1)}_j}{\taus} \right) \right] \,.
\end{equation}

\subsubsection{Calculating causal sets for nLIF}\label{si:causalsets}

To find the causal set, we use the following approach:
In case of an nLIF neuron that has $N_{\ell-1}$ input spike times $t^{(\ell-1)}_j$ with weights $W^{(\ell)}_{ij}$, we first define $\presyn = \{j_1, j_2, ..., j_{N_{\ell-1}}\}$ with $t_{j_1} \leq t_{j_2} \leq ... \leq t_{j_{N_{\ell-1}}}$.
Furthermore, we set $\presyn_k = \{j_1, ..., j_k\}$ for $k > 0$.
The causal set is then given by the subset $\presyn_m$ with the smallest index $m$ satisfying
\begin{equation}\nonumber
   \textbf{1.}\ \sum_{j \in \presyn_m} W^{(\ell)}_{ij} \geq \thresh \ \ \ \ \textbf{and 2.} \ \ 
    \presyn_m = \{j: t^{(\ell-1)}_j \leq t^{(\ell)}_i\}\,, t^{(\ell)}_i = \taus \ln \bigg(  \frac{\sum_{j \in \presyn_m} W^{(\ell)}_{ij} e^{t^{(\ell-1)}_j /\ \taus}}{\sum_{j \in \presyn_m} W^{(\ell)}_{ij} - \thresh} \bigg) \,.
\end{equation}
These two conditions are summarised as follows: (1) the inputs have to be strong enough to drive the membrane potential across the threshold, and (2) all inputs that did not cause the spike at time $t^{(\ell)}_i$ occur after it.
The criterion of selecting the set with minimal $m$ ensures that we find the earliest possible output spike time.
If no such set is found, the causal set is defined as the empty set, reflecting the fact that none of the inputs caused the neuron to spike.
In simulations, we set the output spike time to a sufficiently large value such that it does not affect any other neuron in the network, emulating spiking at infinity.

\subsection{Additional theorems}

\subsubsection{Lipschitz continuity}\label{si:theoremLipschitz}
To ease the notation, we drop the nested list notation of causal subnetworks in the following.
We first state the result for a single nLIF neuron (\cref{theorem:lipschitz}).
\vspace{-2.pt}
\begin{challengebox}
\begin{theorem}[Lipschitz continuous]\label{theorem:lipschitz}
    Let $N_0\in \mathbb{N}$, $j \in [1, N_0]$, and $\causal^{(1)}_1 \subset [1, \dots, N_0]$. Moreover, let $a,b \in \mathbb{P}[\causal^{(1)}_1]$ be the input to a single nLIF neuron with $N_0$ input times. 
    Then the output spike time (\cref{eq:spiketime}) is Lipschitz continuous with respect to input times and weights $W^{1}_{1j} \in \mathbb{R}$, $j \in [1, N_0]$:
    \begin{equation}
    \left\| t^{(1)}_1(a) - t^{(1)}_1(b) \right\|_{L^\infty(\mathbb{P}[\mathcal{C}^{(1)}_1])} \leq 2|\causal^{(1)}_1| \mathrm{max}\left(\frac{\bar{W}}{\delta}, \frac{\taus}{\delta}\right) \left\|a - b\right\|_{L^\infty(\mathbb{P}[\mathcal{C}^{(1)}_1])} \,,
\end{equation}
    where  $|\causal|$ denotes the cardinality of $\causal$, $\| W^{(1)}_{1j} \| \leq \bar{W}$, $\delta < \sum_{j \in \causal^{(1)}_1} W^{(1)}_{1j} - \thresh$.
\end{theorem}
\end{challengebox}
\vspace{-2pt}
The proof is given in \cref{si:continuous,si:lipschitz}. 
In addition, the output spike time may change continuously, discontinuously, or become undefined when transitioning between causal pieces.
Which of these occurs can be determined by inspecting the causal set: if all added or removed input neurons have identical spike times, the output spike time changes continuously; otherwise, it changes discontinuously.
If the causal set would reach maximum size, but all inputs together do not reach the threshold, the output spike disappears.
The corresponding result for entire networks follows from the fact that the composition of Lipschitz-continuous functions is itself Lipschitz continuous.

The Lipschitz constant can be reduced by choosing parameters that produce sparsely populated causal sets (small $|\causal|$) that strongly overstep the threshold (large $\delta$).
However, the contribution of the size of the causal sets in the Lipschitz constant is counter-balanced by the maximum weight $\bar {W}$, which has to be increased with decreasing set sizes to ensure that the sum of the weights exceeds $\thresh$.

\subsection{Mathematical proofs}

\subsubsection{Proof of continuity and differentiability}\label{si:continuous}

To improve readability, we drop the layer and output neuron indices in the following. First note that within a causal piece, the output spike time \cref{eq:spiketime} is a composition of continuous and differentiable functions, and hence itself continuous and differentiable with respect to input spike times and weights.

In the following, we prove under which conditions the output spike time is a continuous function of input spike times and weights when crossing between neighbouring causal pieces.
First, let $\causal$ be the causal set of an nLIF neuron with input spike times $[t_0, ..., t_{N-1}]$, weights $[W_0, ..., W_{N-1}]$, and output spike time 
\begin{equation}
    t = \taus \ln \big(T\big) = \taus \ln \bigg(  \frac{\sum_{j \in \causal}W_j e^{t_j /\ \taus}}{\sum_{j \in \causal} W_j - \thresh} \bigg) \,.
\end{equation}
Let $\causal'$ be the causal set of a neighbouring causal piece, with spike times $[\tilde{t}_0, ..., \tilde{t}_{N-1}, \tilde{t}_N]$, weights $[\tilde{W}_0, ..., \tilde{W}_{N-1}, \tilde{W}_N]$, and output spike time $\tilde{t}$:
\begin{equation}
    \tilde{t} = \taus \ln \big(\tilde{T}\big) = \taus \ln \bigg(  \frac{\sum_{j \in \causal}\tilde{W}_j e^{\tilde{t}_j  /\ \taus} + \tilde{W}_N e^{\tilde{t}_N /\ \taus}}{\sum_{j \in \causal} \tilde{W}_j + \tilde{W}_N - \thresh} \bigg) \,.
\end{equation}
We assume that the output spike time of $\causal$ is along the border between the two causal pieces, meaning that $t = t_N$.
Since output spike times can be shifted by $\Delta$ by shifting all input spike times by $\Delta$, without loss of generality, we assume that $\forall x \in \{t, \tilde{t}, t_0, ..., t_N, \tilde{t}_0, ..., \tilde{t_N} \}$, $x \geq 0$.
All spike times are finite, thus $\exists t_\mathrm{max}$ with $0 < t_\mathrm{max} < \infty$ such that $\forall x \in \{t, \tilde{t}, t_0, ..., t_N, \tilde{t}_0, ..., \tilde{t_N} \}$, $x \leq t_\mathrm{max}$.
Similarly, $\exists \bar{W} > 0$ such that $\forall \omega \in \{W_0, ..., W_N, \tilde{W}_0, ..., \tilde{W}_N\}$, $\| \omega\| \leq \bar{W}$.
Furthermore, $\exists \epsilon_\thresh$ with $0 < \epsilon_\thresh < \infty$ such that $\epsilon_\thresh < \sum_{j \in \causal} \tilde{W}_j + \tilde{W}_N - \thresh$.
Lastly, we highlight the following identity:
\begin{align}
    T &= T \cdot \frac{\sum_{j \in \causal} W_j + M - \thresh}{\sum_{j \in \causal} W_j + M - \thresh} \\
    &= T \cdot \frac{\sum_{j \in \causal} W_j - \thresh}{\sum_{j \in \causal} W_j + M - \thresh} +  \frac{M \cdot T}{\sum_{j \in \causal} W_j + M - \thresh} \\
    &= \frac{\sum_{j \in \causal}W_j e^{t_j /\ \taus}}{\sum_{j \in \causal} W_j - \thresh} \cdot \frac{\sum_{j \in \causal} W_j - \thresh}{\sum_{j \in \causal} W_j + M - \thresh} +  \frac{M \cdot T}{\sum_{j \in \causal} W_j + M - \thresh} \\
    &= \frac{\sum_{j \in \causal}W_j e^{t_j /\ \taus} + M \cdot e^{t_N /\ \taus}}{\sum_{j \in \causal} W_j + M - \thresh} \label{eqsi:identity}
\end{align}
for all $M \in \mathbb{R}$ with $\sum_{j \in \causal} W_j + M - \thresh > 0$.

We first prove continuity for the argument of the logarithm by showing that $\forall \epsilon > 0$, $\exists \delta > 0$ such that $\|t_j - \tilde{t}_j \| < \delta$ with $j \in [0, N]$, $\|W_j - \tilde{W}_j \| < \delta$ with $j \in [0, N-1]$\footnote{Note that $W_N$ and $\tilde{W}_N$ cannot cause a switch between the two causal sets.}, and $\| T - \tilde{T}\| < \epsilon$.
Using \cref{eqsi:identity}, we have:
\begin{align}
    \ &\| T - \tilde{T}\| \\
    %%%%
    =\ &\bigg\| \frac{\sum_{j \in \causal} W_j e^{t_j /\ \taus} + \sum_{j \in \causal} \tilde{W}_j e^{t_N /\ \taus} + \tilde{W}_N e^{t_N /\ \taus} - \sum_{j \in \causal} W_j e^{t_N /\ \taus}}{\sum_{j \in \causal} \tilde{W}_j + \tilde{W}_N - \thresh} \nonumber \\
    \ \ \ \ \ \ & \frac{ - \sum_{j \in \causal}\tilde{W}_j e^{\tilde{t}_j  /\ \taus} - \tilde{W}_N e^{\tilde{t}_N /\ \taus}}{\sum_{j \in \causal} \tilde{W}_j + \tilde{W}_N - \thresh} \bigg\| \\
    %%%%
    \leq\ &\frac{1}{\epsilon_\thresh}\bigg(
    \|\tilde{W}_N\|\cdot \|e^{\tilde{t}_N /\ \taus} - e^{t_N /\ \taus}\| 
    +\sum_{j \in \causal}\| W_j \| \cdot \|e^{t_j /\ \taus} - e^{\tilde{t}_j /\ \taus}\| \nonumber \\
    \ \ \ \ \ \ & + \|W_j - \tilde{W}_j\|\cdot\| e^{\tilde{t}_j /\ \taus} - e^{t_N /\ \taus} \| 
    \bigg) \,.
\end{align}
In the first step, we used \cref{eqsi:identity} with $M = \sum_{j \in \causal} (\tilde{W}_j - W_j) + \tilde{W}_N$, which leads to both $T$ and $\tilde{T}$ having the same denominator.
Furthermore, we added the term $\sum_{j \in \causal} W_j e^{\tilde{t}_j /\ \taus} - \sum_{j \in \causal} W_j e^{\tilde{t}_j /\ \taus}$ in the numerator.
In the next step, we used $\frac{1}{\epsilon_\thresh} \geq \frac{1}{\sum_{j \in \causal} \tilde{W}_j + \tilde{W}_N - \thresh}$, and applied the triangle inequality several times. Using $\| \tilde{W}_j \| \leq \bar{W}$ $\forall j \in [0,N]$, $\| e^{\tilde{t}_j /\ \taus} - e^{t_N /\ \taus} \| \leq \|1 - C \|$ with $C = e^{t_\mathrm{max} /\ \taus}$, and the mean value theorem for the exponential function, we then obtain:
\begin{align}
    \| T - \tilde{T}\| \leq \frac{C}{\epsilon_\thresh \taus} \left(\sum_{j \in \causal'} \bar{W} \|\tilde{t}_j - t_j \| 
    + \sum_{j \in \causal} \frac{\taus \| 1 - C\|}{C} \|\tilde{W}_j - W_j \|\right) \,.
\end{align}
Choosing $\| \tilde{W}_j - W_j \| < \delta_W$ with $\delta_W = \frac{\epsilon_\thresh}{2 N \|1-C\|} \cdot \epsilon$ and $\| \tilde{t}_j - t_j \| < \delta_t$ with $\delta_t = \frac{\epsilon_\thresh \taus}{C \cdot \bar{W} \cdot 2(N+1)}\cdot \epsilon$, we arrive at
\begin{equation}
    \| T - \tilde{T}\| < \epsilon \,.
\end{equation}
The proof concludes by setting $\delta = \mathrm{min}(\delta_W, \delta_t)$. Continuity of the spike times then follows from the fact that the concatenation of continuous functions is again a continuous function. 

Here we assumed that the neighbouring causal set $\causal'$ has the property $\sum_{j \in \causal'} \tilde{W}_j - \thresh > 0$. If this is not the case, then at least one more input neuron with spike time $t^* = \mathrm{min}_x\{t_x \ \ |\ \ x \in \presyn\, \backslash \, \causal'\}$ (with $t^* > t)$ has to be added to the causal set until the condition holds again. Since the new output spike time has to be larger than $t^*$, its value jumps and is therefore not continuous when passing between causal pieces.

\subsubsection{Lipschitz constants} \label{si:lipschitz}

To improve readability, we drop the layer and output neuron indices in the following. Within a causal piece $\causal$, the causal set does not change and the output spike time $t^*$ (\cref{eq:spiketime}) is a composition of continuous and differentiable functions, and is therefore also continuous and differentiable.
Hence, we estimate the Lipschitz constant by bounding the first derivative of the output spike time $t^*$.

Let $\causal$ be a causal set with corresponding input spike times $t_0, ..., t_{N-1}$ for $N \in \mathbb{N}$, weights $W_0, ..., W_{N-1}$, and output spike time $t^*$.
As in the previous subsection, we assume an upper bound for the absolute value of the weights, i.e., $\exists \bar{W} >0$ such that $\forall \omega \in \{W_0, ..., W_{N-1}\}$, $\| \omega \| \leq \bar{W}$. 
Moreover, we assume that all spike times are larger or equal to $0$, and we choose a $\delta > 0$ such that $\delta \leq \sum_j W_j - \thresh$.

We first calculate the Lipschitz constant with respect to input spike times:
\begin{align}
    \left\| \pderivs{t^*}{t_k} \right\| 
    &= \left\|\pderivs{}{t_k} \taus \mathrm{ln} \left( \frac{\sum_{j \in \causal} W_j e^{t_j /\ \taus}}{\sum_{j \in \causal} W_j - \thresh} \right) \right\| \\
    &= e^{-t^* /\ \taus} \left\| \frac{W_k e^{t_k /\ \taus}}{\sum_j W_j - \thresh} \right\| \\
    &\leq \frac{\bar{W}}{\delta} \,,  
\end{align}
where we used that $e^{(t_k - t^*) /\ \taus} \leq 1$ since $t^* \geq t_k$ by definition.

For weights, we get:

\begin{align}
    \left\| \pderivs{t^*}{W_k} \right\| 
    &= \left\|\pderivs{}{W_k} \taus \mathrm{ln} \left( \frac{\sum_{j \in \causal} W_j e^{t_j /\ \taus}}{\sum_{j \in \causal} W_j - \thresh} \right) \right\| \\
    &= \taus e^{-t^* /\ \taus} \left\| \frac{e^{t_k /\ \taus}}{\sum_j W_j - \thresh} - \frac{\sum_{j \in \causal}W_j e^{t_j /\ \taus}}{(\sum_j W_j - \thresh)^2} \right\| \\
     &= \taus e^{-t^* /\ \taus} \left\| \frac{e^{t_k /\ \taus} - e^{t^* /\ \taus}}{\sum_j W_j - \thresh} \right\| \\
     &= \taus \left\| \frac{e^{(t_k - t^*) /\ \taus} - 1}{\sum_j W_j - \thresh} \right\| \\
    &\leq \frac{\taus}{\delta} \,,  
\end{align}
where we used that $0 \leq e^{(t_k - t^*) /\ \taus} \leq 1$ by definition, and hence $\|e^{(t_k - t^*) /\ \taus} - 1\| \leq 1$.

Thus, for a causal piece $\mathbb{P}_\mathcal{C} \subseteq \mathbb{R}^{d \times d}$, where $d \in \mathbb{N}$ is the dimension of the input, and $a,b \in \mathbb{P}_\mathcal{C}$ we have:
\begin{equation}
    \left\| t(a) - t(b) \right\|_{L^\infty(\mathbb{P}_\mathcal{C})} \leq 2|\causal| \mathrm{max}\left(\frac{\bar{W}}{\delta}, \frac{\taus}{\delta}\right) \left\|a - b\right\|_{L^\infty(\mathbb{P}_\mathcal{C})}
\end{equation}
where $L_{\mathbb{P}_\mathcal{C}} = 2|\causal| \mathrm{max}\left(\frac{\bar{W}}{\delta}, \frac{\taus}{\delta}\right)$ is the Lipschitz constant of causal piece $\mathbb{P}_\mathcal{C}$ with causal set $\causal$, and $|\causal|$ is the number of elements in the causal set.

\subsubsection{Proof of \cref{theorem:nLifbound}}\label{si:proofBound}

To improve readability, we drop the layer indices in the following. First, we recapitulate the following theorem which holds, for example, for ReLU neural networks \cite{frenzen2010number} (Theorem 2)\footnote{See also \cite{petersen2024mathematical}, Theorem 6.2}:
\vspace{-2.pt}
\begin{challengebox}
\begin{theorem}\label{lemma:reluapprox}
Let $-\infty < a < b <\infty$, $f \in C^3([a,b])$ and $f$ is not affine. Then there exists a constant $c > 0$ that only depends on $\int_a^b \sqrt{|f''(x)|} \mathrm{d}x$ so that
\begin{equation}
    \|\psi - f \|_{L^\infty([a,b])} > c \cdot p^{-2} 
\end{equation}

for all piecewise linear $\psi$ with $p \in \mathbb{N}$ number of linear pieces. 
\end{theorem}
\end{challengebox}
\vspace{-2pt}
\cref{eq:spiketime} can be written as a piecewise linear function by substituting $\subs{i} = e^{\outspikei{i} / \taus}$ \cite{mostafa2017supervised}, leading to:
\begin{align}
    \subs{i} &=  \frac{1}{\weightsumth{i}{j}} \cdot \expsumT{i}{k}\,. \label{eq:linearT}
\end{align}
As input, we provide $t_x = \taus \ln\left(\frac{e^{a/\taus}(b-x) + e^{b/\taus}(x-a)}{b-a}\right)$ with $x \in [a, b]$ and $-\infty < a < b < \infty$. An nLIF neural network $\Psi(x)$ using this substitution is then a composition of piecewise linear functions, and hence also itself a piecewise linear function of $x$. In this case, \cref{lemma:reluapprox} applies to $\Psi$. 
The output of an equivalent nLIF network $\Phi$ without substitution is given by $\Phi = \taus\mathrm{ln}\Psi$, i.e., we only apply the logarithm to the final output and scale by $\taus$. This can be used to derive \cref{theorem:nLifbound}:
\begin{align}
    \| \Phi - g \|_{L^\infty\left([a,b]\right)} &= \taus \left\| \mathrm{ln}\Psi - \mathrm{ln}\left(e^{g /\ \taus}\right) \right\|_{L^\infty\left([a,b]\right)} \,,\\
    % &= \frac{1}{\zeta^*} \| \Psi - e^g \|_{L^\infty\left([a,b]\right)}\,, \ \ \ \ \mathrm{for} \ \ \ \ \zeta^* \in [\Psi, e^g] \,, \label{eqsi:mvt} \\
    &\geq \frac{\taus}{\zeta} \| \Psi - e^{g /\ \taus} \|_{L^\infty\left([a,b]\right)}\,, \ \ \ \ \mathrm{with} \ \ \ \ \zeta = \mathrm{max}\left[\mathrm{max}_x(\Psi(x)), \mathrm{max}_x(e^{g(x) /\ \taus})\right] \,, \label{eqsi:mvt}  \\
    &> \frac{c}{\zeta} p^{-2}\,, \ \ \ \ \mathrm{with} \ \ \ \ c > 0 \ \ \mathrm{depending\ only\ on}\ \ \taus\int_a^b \sqrt{\left|\frac{\mathrm{d}^2}{\mathrm{d}x^2}e^{g(x) /\ \taus}\right|} \mathrm{d}x  \label{eqsi:pieces} \,,
\end{align}
where we applied the mean value theorem to arrive at \cref{eqsi:mvt} (i.e., we apply the mean value theorem to get rid of the logarithms) and \cref{lemma:reluapprox} to arrive at \cref{eqsi:pieces}. For the latter, we used the fact that $e^{g/\ \taus} \in C^3([a,b])$ is not affine, allowing us to apply \cref{lemma:reluapprox} using $f = e^{g /\ \taus}$. Furthermore, we note that $\Phi$ and $\Psi$ have the same number of causal pieces.

\subsubsection{Random walks}\label{si:randomWalk}

We drop the layer and output neuron index notation used in the main text to clear up the notation.
Assume we have a single neuron with $N_0$ inputs. 
Let $\mathcal{K} = \{j_1, ..., j_K\} \subseteq [1, N_0]$ with $1 \leq K \leq N_0$, let $t_j$ be the input times and $W_{j} \in \mathbb{R}$ the corresponding weights, with $j \in [1, N_0]$.
We denote by $p^q_k$ the probability that the subset $\mathcal{K}$ is a causal set if weights $W_j \sim q$ are sampled from a distribution $q$.

For $\mathcal{K}$ to be a causal set, we have to check the two conditions mentioned in \cref{sec:method}.
The first condition is satisfied if 
\begin{equation}\label{eq:siWcond}
    \sum_{i \in \mathcal{K}} W_i \geq \thresh \,.   
\end{equation}
Assuming the weights are sampled from a random distribution, this can be viewed as a random walk with discrete steps and randomly sampled, continuous step sizes.
The position of the random walk at step $k$ is given by $S_{k} = \sum_{i=1}^{k} W_i$.
In this framework, the first condition becomes the question of whether the random walk is above or equal to the threshold at step $K$, i.e., $S_K \geq \thresh$.

The second condition -- only spike times belonging to the causal set appearing before the output spike -- can always be achieved by choosing inputs the following way (this does not apply to deep networks):
\begin{enumerate}
    \item Set  $t_{j} = c$ for $c \in \mathbb{R}$ and $j \in \{j_\ell, ..., j_K\}$.
    \item Since condition 1 is satisfied, use \cref{eq:spiketime} to calculate the output spike time $t$ with $\mathcal{K}$ as the causal set.
    \item Set $t_{j} > t$ for $j \in \{j_\ell, ..., j_K\}$.
\end{enumerate}
This way, any subset that suffices the first condition (sum of weights above threshold) is a valid causal set.
Since we can choose inputs arbitrarily for a single nLIF neuron, $p^q_k$ is identical to the probability of the random walker to be above threshold at step $k$.

The values of $p^q_k$ are lower bounded by the first-passage-time distribution of the random walk.
That's because the number of trajectories being above or equal to the threshold at step $k$ is lower-bounded by the number of trajectories that cross the threshold for the first time at step $k$.

\subsubsection{Proof of \cref{theorem:lbound}}\label{si:proofSparre}

Let $N \in \mathbb{N}$ be the number of inputs of a single nLIF neuron.
We define $S_{n} = \sum_{i=1}^{n} W_i$ as the cumulative sum of weights $W_i \in \mathbb{R}$ with $S_0 = 0$ and $0 \leq n \leq N$.
For the proof, we first note that $p^{q}_{n} \geq p_\mathrm{FPT}(n)$, where $p_\mathrm{FPT}(n) = p(S_{n} \geq \thresh, S_{n-1} < \thresh, S_{n-2} < \thresh, ..., S_1 < \thresh)$ is the first-passage-time distribution (at step $n$) for a random walk with discrete steps and random continuous step sizes ($W_j \sim q$), see \cref{si:randomWalk}.

In the assumed limit, the survival probability, i.e., not passing the threshold until step $n+1$, is given by the Sparre Andersen theorem \cite{andersen1954fluctuations,majumdar2010universal}:
\begin{equation}
     Q(n) = p( S_{n} < \thresh, S_{n-1} < \thresh, ..., S_1 < \thresh) = \frac{1}{2^{2n}}\binom{2n}{n} \,.
\end{equation}
The first-passage-time probability for step $n+1$ is obtained by taking the difference of survival probabilities:
\begin{align}
    p_\mathrm{FPT}(n+1) &= Q(n) - Q(n+1) \\
    &= \frac{1}{2^{2n}}\binom{2n}{n} - \frac{1}{2^{2n+2}}\binom{2n+2}{n+1} \\
    &= \frac{1}{2^{2n+1}} \binom{2n}{n} \left[2 - \frac{(2n+2)(2n+1)}{2(n+1)(n+1)} \right] \\
    &= \frac{1}{2^{2n+1}} \binom{2n}{n} \left[2 - \frac{(2n+1)}{(n+1)} \right] \\
    &= \frac{1}{2^{2n+1}} \binom{2n}{n} \frac{1}{n+1} \\
    &= \frac{C_n}{2^{2n+1}} \,,
\end{align}
with the Catalan number $C_n = \frac{1}{n+1}\binom{2n}{n}$.
Using a lower bound for the Catalan number \cite{stackoverflowCatalanBounds}, we get:
\begin{equation}
    p^{q}_{n+1} \geq p_\mathrm{FPT}(n+1)  \geq \frac{1}{2 (n+1) \sqrt{\pi \cdot \left(n + \frac{1}{3} \right)}} \,.
\end{equation}
This expression is monotonically decreasing, hence it reaches its minimum value at $n = N-1$:
\begin{equation}
    p^{q}_{n+1} \geq \frac{1}{2 N \sqrt{\pi \cdot \left(N - \frac{2}{3} \right)}} \,.
\end{equation}
Using this, we can estimate the number of causal pieces:
\begin{align}
    \eta^q &= \sum_{k=1}^{N} \binom{N}{k} p^{q}_k \\
    &\geq \sum_{k=1}^{N} \binom{N}{k} p_\mathrm{FPT}(k) \\
    &\geq \frac{1}{2 N \sqrt{\pi \cdot \left(N - \frac{2}{3} \right)}} \cdot \sum_{k=1}^{N} \binom{N}{k} \\
    &= \frac{2^N - 1}{2 N \sqrt{\pi \cdot \left(N - \frac{2}{3} \right)}} \,.
\end{align}

\subsubsection{Number of pieces}\label{si:piecesComb}

For a single nLIF neuron, the number of pieces is obtained combinatorically: given $N$ inputs to the neuron, we can create $\binom{N}{k}$ different subsets with $k$ entries from these neurons. We denote by $p^{q}_{k}$ the probability that, if weights are sampled from a probability distribution $q$, a subset of $k$ inputs forms a causal set. The total number of causal pieces is then obtained by summing up the contributions of subsets of different length: 
\begin{equation}
    \eta = \sum_{k = 1}^N \binom{N}{k} p^q_k \,.
\end{equation}
The upper bound is obtained by using $p^q_k \leq 1$ for all $k$, and therefore $\eta \leq \sum_{k = 1}^N \binom{N}{k} = 2^N - 1$.

For deep networks, we first look at a 2-layer network with $\{N_1, N_2, 1\}$ neurons, where $N_1$ is the number of inputs to the network.
Starting with the output neuron, we can construct a single causal piece as follows:
first, we sample a set of $r$ inputs. From the analysis for single nLIF neurons, we know that $\binom{N_2}{r} p^{q_2}_{r}$ such sets exist.
Next, we have to estimate the number of pieces of the $r$ selected input neurons, which are all given by $\eta_1 = \sum_{k = 1}^{N_1} \binom{N_1}{k} p^{q_1}_k$.
However, the causal piece of the output neuron changes if any of its $r$ selected input neurons change their causal set.
Thus, the number of pieces is given by $\binom{N_2}{r} p^{q_2}_{r} \eta_1^{r}$ -- assuming the best case where the pieces of the output neuron are maximally split up by the input neurons.
The total number is then given by:
\begin{align}
    \eta_2 &= \sum_{r = 1}^{N_2}\binom{N_2}{r} p^{q_2}_{r} \eta_1^{r}\,.
\end{align}
More generally, we have:
\begin{equation}\label{eq:recursive}
    \eta_n = \sum_{r = 1}^{N_{n}}\binom{N_{n}}{r} p^{q_{n}}_{r} \eta_{n-1}^{r}\,,
\end{equation}
for $0 < n \leq \ell$ and $\eta_0 = 1$, where $\ell$ is the number of layers.
Using $p^{q_n}_r \leq 1$ for all $n$ and $r$ and the binomial formula, we get:
\begin{align}
    \eta_{n} &\leq (1+\eta_{n-1})^{N_n} - 1 \,.
\end{align}
Defining $x_n = \eta_n + 1$ leads to $x_{n} \leq x_{n-1}^{N_n}$. Starting with $n=\ell$ until we arrive at $n = 1$, we then get:
\begin{align}
    x_{l} &\leq 2^{\prod_{i=1}^\ell N_i}\\
    &\leq  2^{N^\ell} \,,
\end{align}
with $N = \mathrm{max}\{N_1, N_2, ..., N_\ell, 1\}$. Substituting back, this results in $\eta_l \leq 2^{N^\ell} - 1 < 2^{N^\ell}$.

\subsection{Simulation details}\label{si:details}

In all simulations using nLIF neurons, we use $\taus = 0.5$ and $\thresh = 1$ unless stated otherwise. For LIF neurons, our standard parameters are $\taum = 0.5$ and $\taus = 0.25$.
To implement deep learning models, we used pyTorch \cite{paszke2019pytorch}.
If a neuron does not spike, we artificially set its output spike time to a very late time that does not affect other neurons.
In this work, we use $t_\text{late} = 5$.
Furthermore, linear readout layers ignore any spike times $\geq t_\text{late}$, i.e., outputs are not affected by dead neurons.

Inputs are presented as single spike times to the neural network.
In case of image data, we translate pixel values to spike times as follows: first, we rescale the pixel values to $[0,1]$.
Each data sample $x_i$ is then translated into a spike time latency $1-x_i$. 
For example, in case of MNIST and Fashion MNIST, this way black pixels are mapped to early spike times, while white pixels are mapped to late spike times.

Simulations were run on the ASC-5 infrastructure.
Larger simulations were run on A40 GPUs and AMD Zen3 CPUs (e.g., Fashion-MNIST and EuroSAT), and smaller ones on a NVIDIA Geforce RTX 4060 Laptop GPU (e.g., Yin-Yang).
Individual simulations for Yin-Yang are quite short, taking on the order of minutes, while simulations involving Fashion-MNIST or EuroSAT take a few hours.

\subsubsection{Optimising initialisations}\label{si:evo}

To find optimised initialisation schemes, we use a simple evolutionary method:  
Starting with a list with four different sets for the initial parameters, $P \in \mathbb{R}^{4 \times 4}$, we perturb each set by adding a random value sampled from a normal distribution $\mathcal{N}(0, 0.1^2)$.
We then use all eight sets of parameters to initialise nLIF neural networks with weights sampled from our chosen distribution (e.g., normal, lognormal, uniform).
For each network, we use the Yin-Yang dataset (or any other method) to estimate the number of pieces.
In this case, we sample the input space using a grid ($x \in [0,1], $y$ \in [0,1]$, 100 increments per dimension, constrained to the circular area).
We then take the parameters that produced the four networks with the highest number of pieces and repeat this process, i.e., with using this new list as $P$.
We stop if the number of pieces does not improve after $n \in \mathbb{N}$ loops.

For positive weights, we initialise weights using a lognormal distribution with mean $\alpha_0 \cdot n_{\ell-1}^{-\alpha_1}$ and standard deviation $\alpha_2 \cdot n_{\ell-1}^{-\alpha_3}$, or a uniform distribution $\mathcal{U}(v_0, v_0 + v_1)$ with $v_0 = \beta_0 \cdot n_{\ell-1}^{-\beta_1}$ and $v_1 =  \beta_2 \cdot n_{\ell-1}^{-\beta_3}$.
$n_{\ell-1}$ is the number of neuron projecting into layer $\ell$.
Through the above optimisation loop, we found $\alpha_0 = 1.29$, $\alpha_1 = 0.57$, $\alpha_2 = 0.85$, $\alpha_3 = 0.76$ and $\beta_0 = 0.70$, $\beta_1 = 0.25$, $\beta_2 = 0.80$, $\beta_3 = 0.47$.
The final parameters for normal and uniform (with positive and negative values) are provided in the main text.

\subsubsection{Details: \cref{fig:Intro}}

To initialise the networks, we use a normal distribution with the parameters found using evolutionary optimisation (see main text and \cref{si:evo}).

In panel A and C (left), the causal pieces of the output neuron of a network with $[40, 1]$ neurons is shown (both are the same image, but the one in C is a cut-out).
We sample three random vectors $d_0 \sim \mathcal{N}(0, 1)^{40}$, $d_1 \sim \mathcal{N}(0, 1)^{40}$, $o \sim \mathcal{N}(0, 1)^{40}$.
The inputs $I$ are then obtained by spanning the plane using $I(\alpha_0, \alpha_1) = o + \alpha_0 \cdot (d_0 - o) + \alpha_1 \cdot (d_1 - o)$.
We use $\alpha_0 \in [0,1]$ and $\alpha_1 \in [0,1]$ and $400$ increments per variable.
% To get the line plot, we set $\alpha_1 = 0$ and increase $\alpha_0$ from $0$ to $1$ in $2000$ increments.

For panel C (right), we use the same setup, but for a network with $[40, 40, 1]$ neurons.

\subsubsection{Details: \cref{fig:RW}}

To obtain the results, we used \cref{alg:mc} (see \cref{alg:mcChap}) to estimate the number of pieces of a single nLIF neuron with weights sampled from $\mathcal{N}\left(\mu, \sigma^2\right)$.
We ran the algorithm for values of $\mu$ and $\sigma$ ranging from $0$ to $0.1$ with increments $0.001$.
The maximum number of inputs was set to $100$.
For each initialisation, we sampled $10^4$ weight vectors (per $k$) to estimate $p^q_k$.

\subsubsection{Details: \cref{fig:PiecesVsAcc}}\label{si:fig3details}

The architecture of our shallow network is: 4 spiking input neurons, 30 spiking hidden neurons, and a non-spiking linear layer with 3 output neurons. 
Causal pieces are calculated for the hidden layer.

We perform two scans: one where we ensure that the initial number of pieces is $< 200$ and $> 1$ (scenario A), and where we ensure that the initial number is $> 200$ (scenario B).
This is done to ensure that we have sufficiently many SNNs with low and high piece counts.
For both scenarios, we perform 100 random experiments.
In scenario A, the normal distributions used to initialise the nLIF neural networks, the mean and standard deviation were both sampled from a uniform distribution $\mathcal{U}(-\alpha, 1-\alpha)$ with $\alpha \in [0.1, 0.2, 0.3, 0.4]$ and $\mathcal{U}(0, 1)$, respectively.
In scenario B, we use $\mathcal{U}(-\alpha, 1-\alpha)$ with $\alpha \in [0.1, 0.2, 0.3]$ and $\mathcal{U}(0, 1)$, respectively.
For both scenarios, we continue sampling mean and standard deviation until the initial number of causal pieces is in the required range.

Each reported data point corresponds to one sampled distribution.
We calculate the number of causal pieces using only the $5000$ training samples.
We used the data samples to create the causal piece plots (panels F and G; note that disconnected clusters with similar colours are still different causal pieces).
Networks are trained using the Adam optimiser with a learning rate of $10^{-2}$ (no weight decay), batch size of $200$, and $200$ epochs.
The best test accuracy is reported, calculated using $1000$ test samples (default for Yin-Yang).
As a loss function, we use a cross-entropy loss.

\subsubsection{Details: \cref{fig:fashion}}

For Fashion-MNIST, the trained networks have $[28 \times 28, 200, 10]$ neurons, with the last layer being a standard linear readout layer.

We perform two scans: one where we ensure that the initial number of pieces is $< 4\cdot 10^4$ and $> 1$ (scenario A), and where we ensure that the initial number is $> 4 \cdot 10^4$ (scenario B).
For both scenarios, we perform 50 random experiments.
In scenario A, the normal distributions used to initialise the nLIF neural networks, the mean was sampled from a uniform distribution $\mathcal{U}(-\alpha, 1-\alpha)$ with $\alpha \in [0.4, 0.5, 0.6, 0.7, 0.8]$ randomly selected. The standard deviation was sampled from $\mathcal{U}(0, 1)$.
In scenario B, we use $\mathcal{U}(-\alpha, 1-\alpha)$ with $\alpha \in [0.3]$ randomly selected. The standard deviation was sampled from $\mathcal{U}(0, 1)$.
Each reported data point corresponds to one sampled distribution.
We calculate the number of causal pieces using only the $60000$ training samples.
Networks are trained using the Adam optimiser with a learning rate of $10^{-3}$ (no weight decay), batch size of $100$, and $400$ epochs.
As a loss function, we use the cross-entropy loss.
The best test accuracy is reported.

The setup for EuroSAT RGB is the same, but we use networks with $[16 \times 16 \times 3, 200, 10]$ neurons.
Moreover, the learning rate is set to $10^{-2}$, and scenario A requires that the number of pieces is $< 2 \cdot 10^4$ and $> 1$, while scenario B requires $> 2 \cdot 10^4$.
In scenario B, we use $\mathcal{U}(-\alpha, 1-\alpha)$ with $\alpha \in [0.1, 0.2, 0.3, 0.4]$ randomly selected.
The resolution of the EuroSAT images is decreased from $32\times32\times3$ to $16\times 16 \times 3$, with no significant decrease in the highest achieved test accuracy.

All experiments for LIF are done similarly as for nLIF (for Yin-Yang, see \cref{si:fig3details}).
Only for the nLIF experiments using Fashion-MNIST, we further ran 50 more simulations using the same initialisation scheme as in scenario A, filtering for SNNs with an initial piece count $> 1.8 \cdot 10^4$ and $< 5 \cdot 10^4$.

\subsubsection{Details: \cref{fig:Deep}}

For each data point, we show results of $10$ runs with different random seeds.
To calculate the number of causal pieces, we used an enlarged dataset composed of points obtained from a grid within the data domain, i.e., we evaluated the input space $[0, 1]^2$ using a $400 \times 400$ grid, leading to $124980$ points (only points within the circular area were used).
We obtained qualitatively similar results using a $600 \times 600$ grid.
In panel C, we show the results for a network with $[4, 20, 20, 20, 20, 20, 3]$ and $[4, 40, 40, 40, 40, 40, 3]$ neurons (10 runs with different seeds).
In panel D, the number of pieces of the output layer are shown for lognormal initialisation and (line) shallow networks with $[40, 80, 160, 320, 400$] neurons in the hidden layer, as well as (dotted) deep networks with $[1, 2, 4, 5, 8, 10]$ hidden layers with $40$ neurons each. 
Again, the median over $10$ runs with different random seeds is shown.
For training, the same setup as described in \cref{si:fig3details} was used.

\subsubsection{Details: \cref{fig:PosSNN}}

Networks are initialised by sampling the weights either from a lognormal or uniform distribution, as described in \cref{si:evo}.
To evaluate $p^q_k$, we again use the Monte Carlo approach described in \cref{alg:mcChap}, with a similar setup as in \cref{fig:RW}.
Panel B is created similarly as panel B in \cref{fig:Deep}.
To keep weights $W$ positive, we apply a ReLU function to them in the forward function, $W \mapsto \mathrm{max}(0, W)$. 

For Yin-Yang, we use a network of size $[4, 30, 3]$, with the last layer being a standard linear pyTorch layer.
We train the networks using a batch size of $100$, learning rate of $10^{-3}$, $5000$ epochs, and Adam optimiser without weight decay.
The reference values ($0.638$ and $0.976$) are taken from \cite{kriener2022yin} (best value also for $[4, 30, 3]$ neurons). They further report an accuracy of $0.855$ if only the upper layer is trained, which is also lower than the performance reached by our networks.

For MNIST, we use a network of size $[28\times 28, 200, 100, 10]$, again with the last layer being a standard linear pyTorch layer.
Pixel values are re-scaled to be in the range $[0, 1]$.
Images are flattened and no image transformations are used during training.
We train the networks using a batch size of $100$, learning rate of $10^{-3}$, $200$ epochs, and Adam optimiser without weight decay.
The best performance ($0.9833$) is taken from \cite{kim2024ideal}. For the performance of a linear layer, we show $0.9277$, as, e.g., reported in \cite{senn2024neuronal}.

For EuroSAT, we use a network of size $[16\times 16 \times 3, 200, 100, 10]$, again with the last layer being a standard linear pyTorch layer.
Images are re-scaled to $16 \times 16 \times 3$ (3 colour channels), with pixel values re-scaled to be in the range $[0,1]$.
Furthermore, we apply random horizontal and vertical flips during training.
Images are flattened before they are provided as input to the neural networks.
We train the networks using a batch size of $100$, learning rate of $10^{-2}$, $1000$ epochs, and Adam optimiser without weight decay.
We found that the best performance of an MLP is similar to the one reached by random forests, which is $0.70$.
For the performance of linear models, we use the results achieved using logistic regression ($0.40$).
We also reached $0.34$ using nearest neighbor and $0.47$ using decision trees.

\subsubsection{Algorithms: Monte Carlo approach}\label{alg:mcChap}

In simulations, we use \cref{alg:mc} to calculate $p^q_k$, from which we calculate the improved upper bound derived in the main paper.
A similar algorithm can be used to estimate $p^q_k$ for a static weight vector (with unknown distribution $q$) by randomly sampling subsets from the vector (e.g., in case of the weights in a trained neural network).

\begin{algorithm}
    \caption{Monte Carlo estimate for perceptron}
    \label{alg:mc}
    \begin{algorithmic}[1]
        \Require Distribution $q$, number of samples $num\_samples$, number of inputs $num\_inputs$, threshold $\thresh$
        \State $prob\_set \gets$ list of length $num\_inputs$ filled with $0$.
        \Comment Probability that subset is a causal set.
        \For{$causal\_set\_length = 1$ to $num\_inputs$}
            \For{$sample\_ID = 1$ to $num\_samples$}
                \State $W \gets$ list of length $causal\_set\_length$ with values sampled from $q$
                \State $c \gets causal\_set\_length - 1$ 
                \State $strong\_enough \gets \sum^{c}_{i=0} W_i \geq \thresh$
                \If{$strong\_enough \text{ is } True$}
                    \State $prob\_set[causal\_set\_length] \gets prob\_set[causal\_set\_length]+1$
                \EndIf
            \EndFor
            \State  $prob\_set[causal\_set\_length] \gets prob\_set[causal\_set\_length]\ /\ num\_samples$
        \EndFor        
        \State \Return $prob\_set$
    \end{algorithmic}
\end{algorithm}

\subsubsection{Algorithms: counting pieces}\label{alg:pieces}

\cref{alg:causal_sets_to_IDs} is used to count the number of causal pieces for (i) neurons in a deep neural network, and (ii) per layer.
To count the pieces, we start from the first layer and index the causal sets.
For neurons in the first layer, the causal sets are just composed of the inputs that caused the spike ((\cref{alg:process_causal_set}, line 5).
Each neuron's piece is given by the index we assign it (\cref{alg:assign_id}).
For neurons in deep layers, the causal set consists of both the indices of the inputs that caused it to spike, and the causal piece indices of these neurons (\cref{alg:process_causal_set}, line 3).
For layers (\cref{alg:get_layer_indices}), the causal set is given by the list of causal piece indices of all neurons in the layer.
If any of these indices changes, the causal piece of the layer changes.

\begin{algorithm}[h!]
    \caption{Transform causal sets (per neuron) to causal piece IDs}
    \label{alg:causal_sets_to_IDs}
    \begin{algorithmic}[1]
        \Require Nested list with ordered causal sets, \texttt{sets}. Dimensions are: samples, layers, neurons.
        \State  $causal\_set\_to\_ID \gets \text{empty dictionary}$
        \State  $causal\_set\_to\_ID[\text{String}([])] \gets -1$
        \State $num\_samples \gets \text{length}(sets)$
        \State $IDs \gets \text{list containing } num\_samples \text{ empty lists}$
        \For{$sample\_id = 0$ to $num\_samples - 1$} \Comment{Iterate over samples}
            \State $sets\_of\_sample \gets sets[sample\_id]$
            \For{$layer\_id = 0$ to \text{length}($sets\_of\_sample)-1$} \Comment{Iterate over layers}
                \State $layers \gets sets\_of\_sample[layer\_id]$
                \State Append empty list to $IDs[sample\_id]$
                \For{each $causal\_set$ in $layers$} \Comment{Turn causal set of every neuron to corresponding ID}
                    \State $cset\_name \gets \textsc{ProcessCausalSet}(causal\_set, IDs, sample\_id, layer\_id)$
                    \State $single\_ID \gets \textsc{AssignID}(cset\_name, causal\_set\_to\_ID)$
                    \State Append $single\_ID$ to $IDs[sample\_id][layer\_id]$
                \EndFor
            \EndFor
        \EndFor
        \State \Return $IDs$
    \end{algorithmic}
\end{algorithm}

\begin{algorithm}
    \caption{\textsc{ProcessCausalSet}}
    \label{alg:process_causal_set}
    \begin{algorithmic}[1]
        \Require Causal set $causal\_set$, List of causal set IDs $IDs$, Sample index $sample\_id$, Layer index $layer\_id$
        \If{$layer\_id > 0$}
            \State $prev\_layer\_IDs \gets IDs[sample\_id][layer\_id-1]$
            \State $selected\_IDs \gets$ empty list
            \For{$index$ in $causal\_set$} \Comment{Iterate over indices in causal set}
                \State Append $prev\_layer\_IDs[index]$ to $selected\_IDs$
            \EndFor
            \State $cset\_name \gets \text{String}([selected\_IDs, causal\_set])$
        \Else
            \State $cset\_name \gets \text{String}(causal\_set)$
        \EndIf
        \If{$\text{length}(causal\_set) = 0$}
            \State $cset\_name \gets \text{String}([])$
        \EndIf
        \State \Return $cset\_name$
    \end{algorithmic}
\end{algorithm}

\begin{algorithm}
    \caption{\textsc{AssignID}}
    \label{alg:assign_id}
    \begin{algorithmic}[1]
        \Require Causal set name $cset\_name$, Dictionary $causal\_set\_to\_ID$
        \If{$cset\_name \notin \text{keys}(causal\_set\_to\_ID)$}
            \State $causal\_set\_to\_ID[cset\_name] \gets \text{length}(causal\_set\_to\_ID)$ \Comment{For efficiency, one can also hash $cset\_name$.}
        \EndIf
        \State \Return $causal\_set\_to\_ID[cset\_name]$
    \end{algorithmic}
\end{algorithm}

\begin{algorithm}
    \caption{Get Causal Piece ID for Neural Network Layers}
    \label{alg:get_layer_indices}
    \begin{algorithmic}[1]
        \Require \texttt{IDs}, List of dictionaries \texttt{layer\_indices\_dict} with length $num\_layers -1$
        \State $piece\_ID\_layers \gets$ empty list
        
        \For{$sample\_ID = 0$ to $\text{length}(IDs) - 1$} \Comment{Iterate over samples}
            \State Append empty list to $piece\_ID\_layers$
            \For{$layer\_ID = 0$ to $\text{length}(IDs[sample\_ID]) - 1$} \Comment{Iterate over layers}
                \State $lay\_state \gets \text{String}(IDs[sample\_ID][layer\_ID])$
                \If{$lay\_state \notin \text{keys}(layer\_indices\_dict[layer\_ID])$}
                    \State $layer\_indices\_dict[layer\_ID][lay\_state] \gets \text{length}(layer\_indices\_dict[layer\_ID])$
                \EndIf
                \State Append $layer\_indices\_dict[layer\_ID][lay\_state]$ to $piece\_ID\_layers[sample\_ID]$
            \EndFor
        \EndFor
        
        \State \Return $piece\_ID\_layers$
    \end{algorithmic}
\end{algorithm}

%%%%%%%%%%%%%%%%%%%%%%%%%%%%%%%%%%%%%%%%%%%%%%%%%%%%%%%%%%%%

% \newpage
% \input{checklist.tex}

\end{document}